\documentclass[journal]{IEEEtran}
\ifCLASSINFOpdf
\else
\fi

\usepackage{graphicx}
\usepackage{subfigure}
\usepackage{amsmath,amssymb} 
\usepackage{url}
\usepackage{cite}
\usepackage{color}
\usepackage{caption} 
\usepackage[table]{xcolor}

\begin{document}
%
\title{Image Inpainting by {\color{black}{End-to-End}} Cascaded Refinement with Mask Awareness}
%
%
%

\author{Manyu Zhu,
        Dongliang He,
        Xin Li,
        Chao Li,\\
        Fu Li,
        Xiao Liu,
        Errui Ding,
        Zhaoxiang Zhang,~\IEEEmembership{Senior Member,~IEEE}
\thanks{When this work was done, Manyu Zhu, Dongliang He, Xin Li, Chao Li, Fu Li, Xiao Liu and Errui Ding were all with Department of Computer Vision (VIS) Technology, Baidu Inc. Zhaoxiang Zhang is currently with Institute of Automation, Chinese Academy of Sciences (CASIA), Beijing, China. Corresponding author: Dongliang He (hedlcc@126.com).}}

%
%

\markboth{Journal of IEEE Transactions on Image Processing}
{Shell \MakeLowercase{\textit{et al.}}: Bare Demo of IEEEtran.cls for IEEE Journals}
%



\maketitle

\begin{abstract}
Inpainting arbitrary missing regions is challenging because learning valid features for various masked regions is nontrivial. Though U-shaped encoder-decoder frameworks have been witnessed to be successful, most of them share a common drawback of mask unawareness in feature extraction because all convolution windows (or regions), including those with various shapes of missing pixels, are treated equally and filtered with fixed learned kernels. To this end, we propose our novel mask-aware inpainting solution. Firstly, a {\color{black}{Mask-Aware Dynamic Filtering}} (MADF) module is designed to effectively learn multi-scale features for missing regions in the encoding phase. Specifically, filters for each convolution window are generated from features of the corresponding region of the mask. The second fold of mask awareness is achieved by adopting {\color{black}{Point-wise Normalization}} (PN) in our decoding phase, considering that statistical natures of features at masked points differentiate from those of unmasked points. The proposed PN can tackle this issue by dynamically assigning point-wise scaling {\color{black}{factor}} and bias. Lastly, our model is designed to be {\color{black}{an}} end-to-end cascaded refinement one. Supervision information such as reconstruction loss, perceptual loss and total variation loss is incrementally leveraged to boost the inpainting results from coarse to fine. Effectiveness of the proposed framework is validated both quantitatively and qualitatively via extensive experiments on three public datasets including Places2, CelebA and {\color{black}{Paris StreetView}}.    
\end{abstract}


%
\IEEEpeerreviewmaketitle

\section{Introduction}
%
%
%
%

\IEEEPARstart{I}{mage} inpainting has been extensively adopted in many real-world applications, such as retouching photos, restoring images and concealing errors.
Specifically, the task focuses on recovering a damaged image with a given mask that indicates the missing regions. It faces great challenge because extracting valid features for the missing regions is nontrivial.
Early methods such as \cite{hays2007scene,barnes2009patchmatch,sun2005image,efros1999texture,criminisi2004region} typically search for the best matched patches based on hand-crafted features to fill in the holes. 
They usually can synthesize visually realistic stationary textures, but fail to generate semantically correct results. 

\begin{figure}[t]
\centering
\subfigure{
\begin{minipage}[b]{0.22\columnwidth}
\includegraphics[width=1\columnwidth]{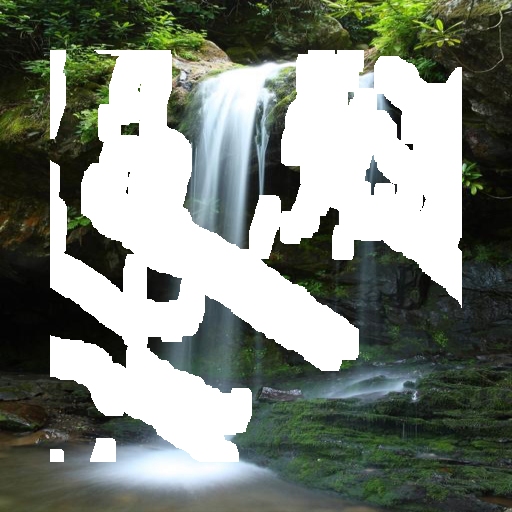}\vspace{1pt}
\includegraphics[width=1\columnwidth]{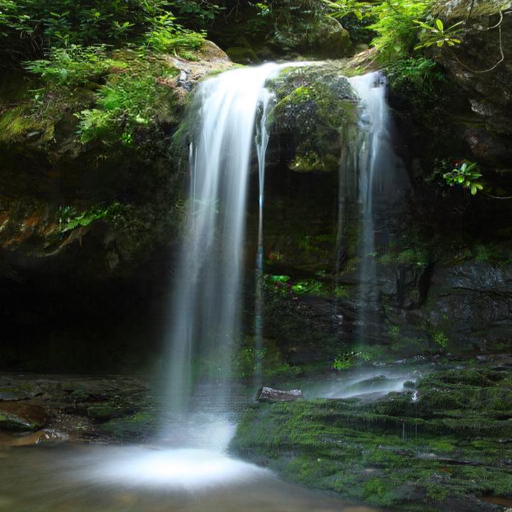}\vspace{1pt}
\end{minipage}}
\subfigure{
\begin{minipage}[b]{0.22\columnwidth}
\includegraphics[width=1\columnwidth]{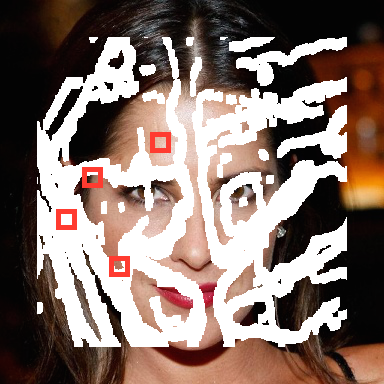}\vspace{1pt}
\includegraphics[width=1\columnwidth]{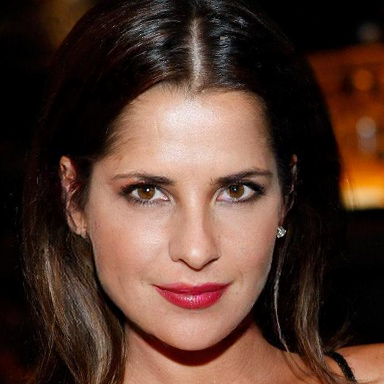}\vspace{1pt}
\end{minipage}}
\subfigure{
\begin{minipage}[b]{0.22\columnwidth}
\includegraphics[width=1\columnwidth]{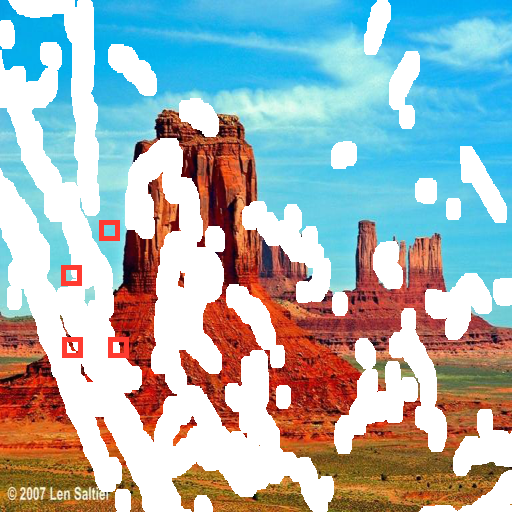}\vspace{1pt}
\includegraphics[width=1\columnwidth]{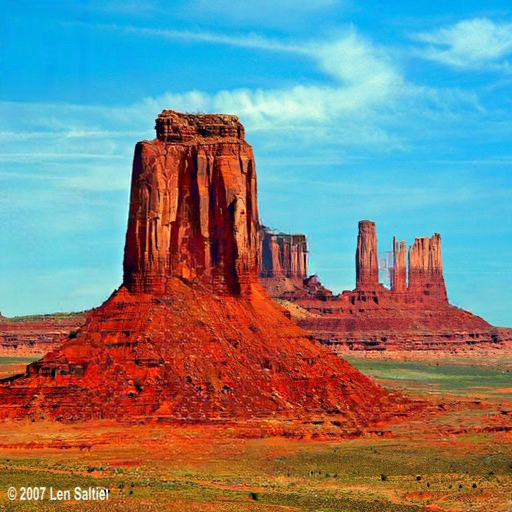}\vspace{1pt}
\end{minipage}}
\subfigure{
\begin{minipage}[b]{0.22\columnwidth}
\includegraphics[width=1\columnwidth]{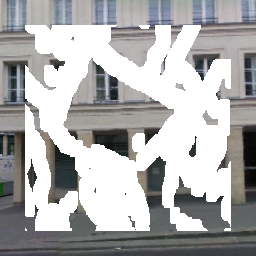}\vspace{1pt}
\includegraphics[width=1\columnwidth]{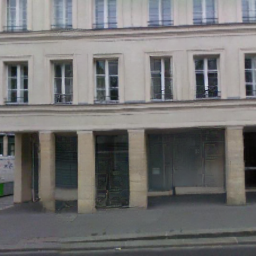}\vspace{1pt}
\end{minipage}}

\caption{Inpainting results generated by our framework. The missing regions are shown in white. The red boxes show local regions with various shaped valid pixels. }
\label{fig:first_show}
\end{figure}
 
Recently, deep encoder-decoder based methods \cite{pathak2016context,iizuka2017globally,yeh2017semantic} become popular for image inpainting. 
The input image is firstly encoded into a latent high-level feature space, and then decoded back to low-level pixels. 
Especially the U-shaped encoder-decoder frameworks show great ability in generating highly structured images. {\color{black}{For instance, PEN-Net \cite{pennet} proposes to progressively learn region affinity by attention and fill holes from low-resolution to high-resolution in a U-shaped pyramid structure.}} There are also two-stage based solutions proposed to synthesize the missing contents in a coarse-to-fine way and to utilize pre-defined intermediate clues to reduce the difficulty of directly generating missing contents. 
For example, \cite{yu2018generative} synthesizes a blurry image as guidance at first for inpainting the missing regions. EdgeConnect \cite{nazeri2019edgeconnect} comprises of an edge generator followed by an image completion network. 
Hallucinated edges of the missing regions are leveraged by image completion network for recovery.
Analogically, SPG-Net \cite{song2018spgnet} factorizes the image inpainting into segmentation repair stage and segmentation guidance stage. Thus image segmentation map is used to guide the pixel generation. {\color{black}{StructureFlow \cite{structureflow} can be regarded as a two-stage solution which firstly reconstructs structure and then generates textures using appearance flow.}}

\begin{figure*}[t]
\centering
\includegraphics[width=0.8\linewidth]{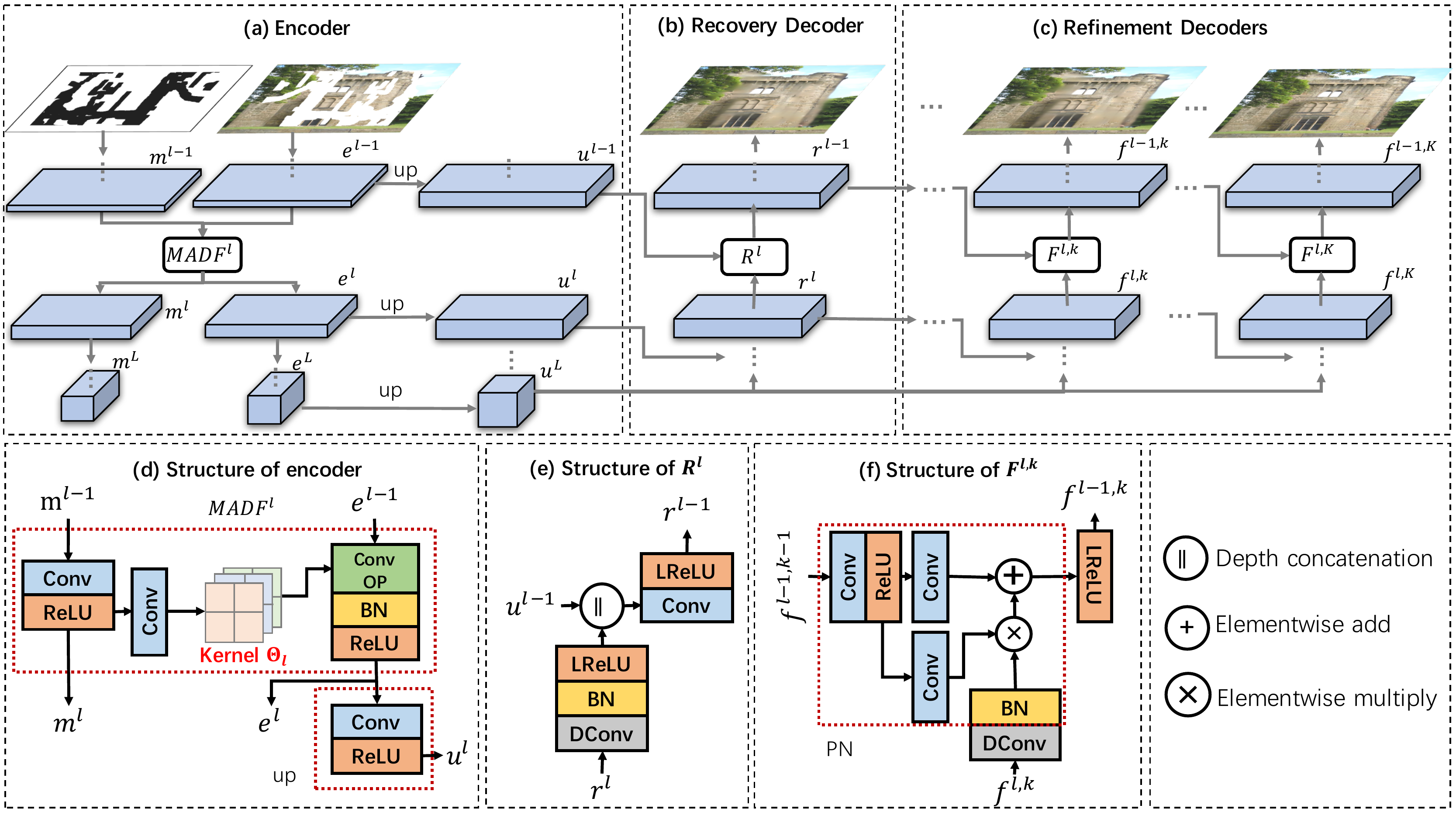}
\caption{The architecture of our framework. $MADF^l$ and $R^l$ are the MADF module and recovery decoder block at $l$-th level respectively. $F^{l,k}$ represents the $k$-th refinement decoder block at $l$-th level. 
$m^l$ is the $l$-th mask feature map of the encoder. The convolution operation marked in green in the encoder takes $e^{l-1}$ as input and its kernel for each convolution window is generated from corresponding region of $m^{l-1}$. 
``DConv'' denotes the transpose convolution, ``LReLU'' denotes the leaky ReLU, and ``up'' denotes increasing channels by $1\times1$ convolution layers. {\color{black}{$u^{l-1}$ and $r^l$ are inputs to $R^l$ to generate the feature map $r^{l-1}$ at the recovery decoder and $F^{l,k}$ takes $f^{l-1, k-1}$ and $f^{l,k}$ as input and generates feature map $f^{l-1,k}$. Note that $r^L$ and $f^{L,k}$ are all equivalent to $u^L$ and $f^{l,1}$ equals $r^l$.}}
}
\label{fig:framework}
\end{figure*}
However, we find existing deep learning based solutions can be further improved from the following aspects. Firstly, conventional convolution operation in the encoding {\color{black}{phase}} is mask unaware. As the red {\color{black}{boxes}} in Figure \ref{fig:first_show} shown, valid pixels in a convolution window are of various shapes due to arbitrary masked regions. 
\textcolor{black}{
It is necessary to adaptively extract features from these regions, however, conventional convolution treats all windows equally, \textit{i.e.}, each of the convolution windows is filtered by fixed kernels. This strategy can hardly handle the variety of shapes of valid pixels.     
}
Secondly, two-stage models usually suffer from artifacts such as blurry boundary and distorted structures when the missing contents are large and complex, because of the lack of necessary cues to infer the missing contents. As indicated by \cite{wang2018image}, if the first stage fails to generate correct intermediate clues, the wrong information would be inherited by the second stage. 
Moreover, the manually defined characteristics {\color{black}{make}} the entire inpainting procedure sensitive to the choice of the intermediate clues.
To address the aforementioned issues, we propose a novel cascaded refinement framework with mask awareness in this paper. 

\textcolor{black}{
The mask awareness is two-fold. Firstly, we propose a {\color{black}{Mask-Aware Dynamic Filtering}} (MADF) module for effectively feature learning in the encoder. In more detail, filters for each convolution window are generated dynamically from features of the corresponding region of the mask. It is noteworthy that, the proposed MADF module is different from gated convolution (GConv) which is proposed in~\cite{yu2019free}. GConv generalizes partial convolution by providing an attention mechanism to select features from output activations. However, the typical convolution operation is unchanged. Once the kernels are learned and fixed, all convolution windows are filtered by the shared kernels. Although GConv is equipped with attention mechanism, secondhand mask information is applied on the output activation through the attention scores, where the vanilla convolutions {\color{black}{stand}} in between. Therefore, the mask information can not be fully exploited in GConv. In contrast, our MADF module dynamically produces customized kernels for each of convolution windows depending on the corresponding mask information. The mask information is utilized directly and adaptively. Hence, our framework is able to handle various shapes of missing regions.
}

Secondly, statistical characteristics (\textit{i.e.}, mean and variance) of features in hole regions could be different from those of non-hole regions at different decoding phases, applying batch normalization using mean and variance calculated from the whole feature map would still suffer from the well-known ``covariant shift'', due to mismatching mean and variance value for both hole and non-hole points. We consider mask awareness from normalization perspective for the first time in image inpainting task. 
We propose to improve batch normalization by adopting {\color{black}{Point-wise Normalization}} (PN) which is achieved by adaptively assigning scaling factor and bias on per feature point basis. Our PN is largely inspired by and closely related to \cite{park2019semantic} and \cite{wang2018recovering} which use semantic clues for spatial feature modulation in image {\color{black}{synthesis}} or super-resolution, aiming at avoiding semantic-layout washing away. We empirically verify similar technique can be utilized for ``covariant shift removal'' in inpainting task.

\textcolor{black}{
The other contribution of our framework is that, the proposed cascaded refinement architecture is capable of stable image inpainting by exploiting the multi-scale feature space in an end-to-end fashion.
Particularly, based on the traditional U-shaped network, we {\color{black}{extend}} the decoder part to recovery decoder and refinement decoders. The refinement decoders are parallel to the recovery decoder. On one hand, the recovery decoder maps the encoded features back to an coarse image. On the other hand, the refinement decoders work parallelly with the recovery decoder. A refinement decoder refines the features from its preceding refinement decoder or the recovery decoder. Furthermore, the inpainting {\color{black}{results}} are progressively refined by incrementally adding supervision information of reconstruction loss, perceptual loss, total variance loss and so on to the recovery decoder and refinement decoders. Thus,  
{\color{black}{our framework relies on none of explicit intermediate clues, what intermediate information is most robust and helpful is learned by end-to-end training. Besides, our model leverages the learned intermediate signals from multi-scale feature spaces as well as the RGB space}}.  
}

Experiments on several publicly available datasets including Places2 \cite{zhou2017places}, CelebA \cite{liu2015deep} and {\color{black}{Paris StreetView}} \cite{doersch2015makes} demonstrate that the proposed framework can generate semantically plausible and richly detailed contents in various scenes even when the missing regions are large and complex. Exemplar results are shown in Figure \ref{fig:first_show}. Comparing with the existing state-of-the-art methods, our approach significantly outperforms them both quantitatively and qualitatively.

{\color{black}{In a nutshell, we summarize our technical contributions in the following:
\begin{itemize}
    \item We firstly propose the ``different kernels for different convolution windows'' mechanism for image inpainting via a novel Mask Aware Dynamic Filtering (MADF) module in the CNN encoder;
    \item Point-wise Normalization (PN) is designed in the CNN decoding phase to avoid ``covariant shift'' issue at hole and non-hole regions introduced by batch normalization;
    \item Our proposed coarse-to-fine cascaded refinement architecture is proven to effectively boost the inpainting performance.
\end{itemize}
}}

\section{Related Work}
{\color{black}{Traditional patch-based image inpainting approaches such as \cite{hays2007scene,sun2005image,le2011examplar,criminisi2004region,texturematching,lee2016laplacianpm} typically propagated appearance information from the remained regions or other source images into the missing regions through various manually defined similarity metrics between patches. Due to the expansive computational costs of patch matching as well as the limitation of their capability, these approaches were not practical in reality. PatchMatch \cite{barnes2009patchmatch}, a fast similar patch searching method had been proposed to accelerate the inpainting process. However, with only low level information, PatchMatch still can not generate semantically correct results especially for complex textures. Partial differential equation (PDE) based methods \cite{bertalmio2000imagepde,grossauer2004combined,bertalmio2006strong} are designed to connect edges or to extend lines also work well in reconstructing lines, curves and small holes, but they suffer from blurring artifacts when completing relatively large regions. There are also geometry based inpainting solutions \cite{le2011examplar,cao2011geometrically} proposed to leverage geometry guidance for image completion based on exemplar. In the literature, statistics based solutions are also extensively studied \cite{levin2003learning,ghorai2019multiple,fadili2009inpainting}. More detailed review of conventional algorithms can be find from \cite{guillemot2013image,buyssens2015exemplar}.}} 
 
Deep learning based method was firstly introduced to image inpainting by Context Encoders \cite{pathak2016context}, which employed an encoder-decoder based structure and trained with adversarial loss \cite{goodfellow2014generative}. Built upon context encoders, Global$\&$Local \cite{iizuka2017globally} employed global and local discriminators to get better consistency around the boundary of missing regions. \cite{yan2018shift} introduced an U-Net structure \cite{ronneberger2015u} combined with special shift-connection layers for filling in holes of arbitrary shape. {\color{black}{PEN-Net \cite{pennet} also leveraged U-Net with attention to progressively reconstruct missing regions. SSDCGN \cite{shen2019single} proposed single-shot dense connected generative network to address the blurry boundary and distorted structure issue of inpainting output to improving performance}}. Some other methods divided image inpainting into two stages, and utilized explicitly defined intermediate clues such as edges/textures \cite{yang2017high,song2017image}, segmentation maps \cite{song2018spgnet}, {\color{black}{image structure \cite{structureflow}}} and blurry images \cite{yu2018generative} as the guidance to synthesize the missing contents. 
\cite{nazeri2019edgeconnect,xiong2019foreground} disentangled the image inpainting problem into the cucontour completion module and image completion module. {\color{black}{There are also frameworks that focus on avoiding the complexity of inpainting model by parallel decoding network \cite{sagong2019pepsi} or concentrate on generating multiple and diverse plausible solutions for image completion \cite{zheng2019pluralistic}.}}
Our model abandons two-stage design to avoid error propagation from the first stage to the final stage and an end-to-end cascaded refinement structure is used to boost the inpainting result. {\color{black}{There are also other frameworks proposed for many other low-level tasks, such as image rain removal \cite{pfderain,drd,wang2020model} and super-resolution \cite{rdn,rcan,dbpn}. Most of them use vanilla convolution and are not easy to be effectively applied to inpainting. Spatial feature transformation \cite{wang2018recovering} was proposed for image super-resolution and it worked quite similar to the gated convolution proposed for image inpainting \cite{yu2019free}.}}

Given convolutional responses are based on both valid pixels and invalid values in the missing holes, these methods would suffer from artifacts such as distortion and blurriness. PConv \cite{liu2018image} proposed partial convolution where the convolution is masked and re-normalized to be based on only valid pixels. \cite{chaohaoLBAM2019} improved the partial convolutional network by a learnable attention map for feature re-normalization and mask updating. \cite{yu2019free} proposed a gated convolution (GConv) which generalized partial convolution by providing a learnable dynamic feature selection mechanism for feature maps at each level. However, in these works the vanilla convolution operation is still unchanged. Once the kernels are learned and fixed, all local convolution windows are treated equally. 
\textcolor{black}{
Different from GConv, where the conventional convolution is improved by re-weighting the output activation, our model applies dynamic convolution filters according to the shape of valid pixels in each local convolution window. Specifically, the proposed MADF module adaptively generates one particular set of kernels for each of the convolution windows and the mask information is exploited during the kernel generating process. Hence, each convolution location has its own one set of kernels which is customized by the MADF module. This approach is more flexible than barely scaling the output activation.  
}

\begin{figure*}[!t]
    \centering
    \includegraphics[width=0.9\textwidth]{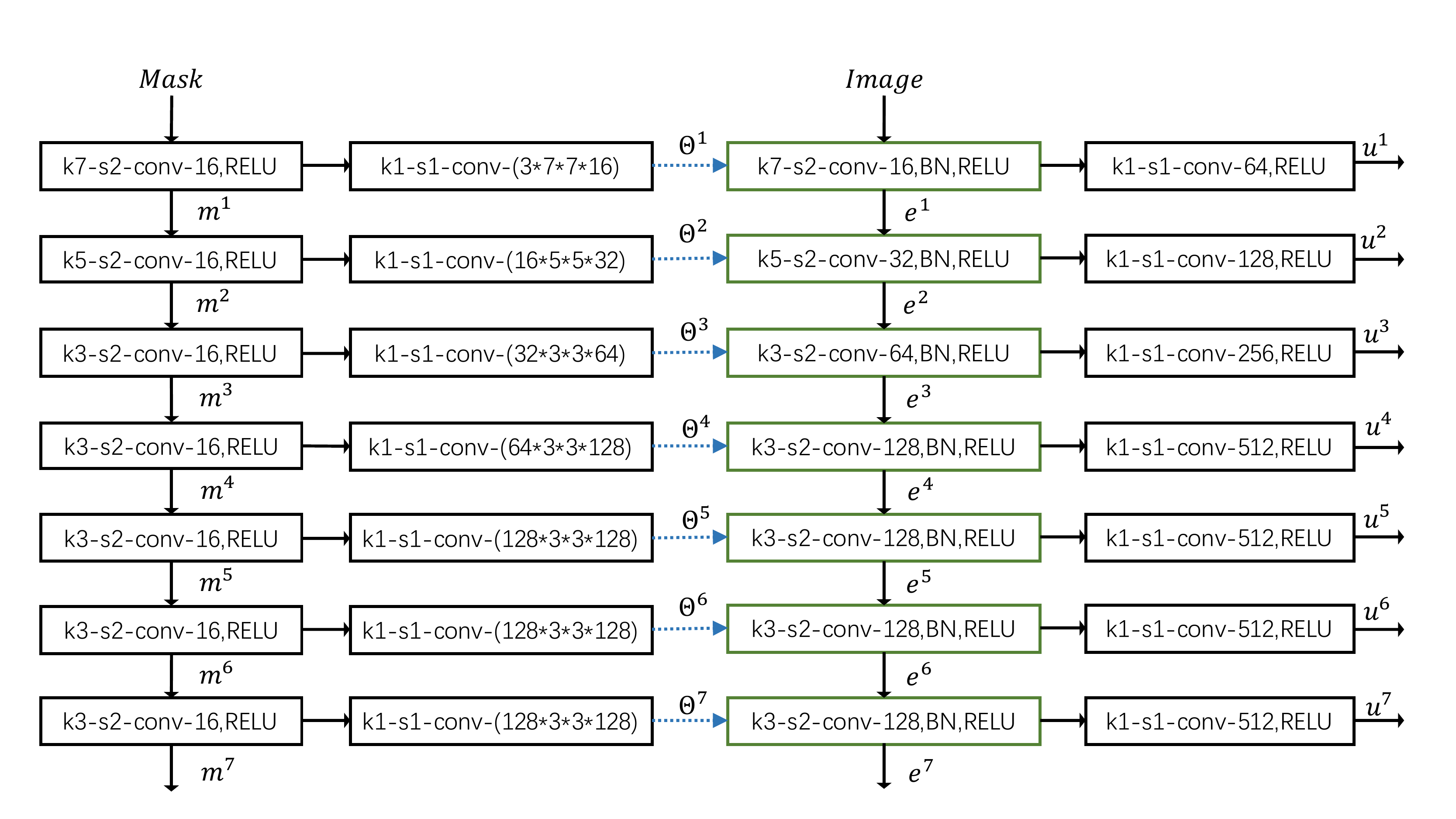}
    \caption{{\color{black}{The details of encoder. ``k*'' denotes kernel size, ``s*'' denotes stride, ``conv-*'' denotes * output channels. The green blocks contain dynamic convolution operations of MADF.}}}
    \label{fig:encoder}
\end{figure*}
\section{Approach}
\subsection{Model Architecture}
Figure \ref{fig:framework} shows the overall architecture of our proposed framework, whose inputs are the damaged image $\mathbf{I_{in}}$ and the corresponding binary mask $\mathbf{M}$. 
Our model consists of three components: a) an encoder $E$ that encodes the damaged image with mask into latent high-level feature maps, b) a recovery decoder $R$ to fill the holes in the feature maps, and c) a sequence of refinement decoders $\{F_1, F_2,\ldots, F_k\}$ to refine the feature maps $K$ times and decode the features back to low-level pixels. We empirically use two refinement decoders to strike a good tradeoff between efficiency and performance. 

{\color{black}{Fisrt of all, we describe the notations used in the following text. $m^l$ and $e^l$ are the mask feature map and the image feature map generated by our Mask-Aware Dynamic Filtering (MADF) module at the $l^{th}$ level of encoder $E$. $u^l$ is the final encoded feature of the $l^{th}$ level of $E$. We use $r^l$ and $f^{l,k}$ to denote feature map at the $l^{th}$ level of the recovery decoder $R^l$ and the $k^{th}$ refinement decoder $F^{l,k}$. There are in total $L$ levels in our model. Note that $r^L$ and $f^{L,k}$ are all equivalent to $u^L$. $f^{l,1}$ equals $r^l$. We use $C_*^l$ to denote number of feature channels of tensor $*$ at level $l$ and $*\in[u,m,r]$.}}

\subsection{Encoder}
During the encoding phase, the encoder $E$ generates multiple level feature maps. 
In order to make a distinction between missing pixels and valid pixels, we introduce Mask-Aware Dynamic Filtering (MADF). It dynamically generates convolution kernels for each convolution window from the features of the corresponding position on the mask, as shown in Figure \ref{fig:framework} (d). Assume a $k\times k$ convolution with stride $s$ will be applied to $e^{l-1}\in R^{H\times W \times C_e^{l-1}}$, and there are $N_H\times N_W$ convolution windows in total. The mask-aware dynamic convolution layer marked in green operates as follows: on $m^{l-1}\in R^{H\times W \times C_m^{l-1}}$, $k\times k$ convolution with stride $s$ and ReLU activation are firstly applied to generate $m^l\in R^{N_H\times N_W\times C_m^l}$, then we utilize $1\times1$ convolution to generate the kernel tensor $\Theta_l\in R^{N_H\times N_W\times D}$, where $D$ equals $C_e^{l-1}\times k \times k\times C_e^l$. Finally, each of all the $N_H\times N_W$ windows in $e^{l-1}$ is convolved using the kernel reshaped from the corresponding point of $\Theta_l$, \textit{i.e.}, the convolution kernel of the $[n_H,n_W]$-th window in $e^{l-1}$ is reshaped from $\Theta_l[n_H,n_W,:]$. 
To keep computational cost in mind, we choose to apply MADF to extract relatively low dimensional latent feature $e^l$ and then map it to higher dimension $u^l$.

\textcolor{black}{  
It is obvious MADF is different from conventional convolution layers which employ ``one set of kernels for all convolution windows''. Our module works as ``one set of kernels per convolution window''. Remarkably, recent state-of-the-art PConv \cite{liu2018image}, which applies one set of kernels for all windows but re-normalizes outputs on per window basis with rule-based scaling factors oriented from the mask and hand-crafted mask updating mechanism, can be viewed as a special case of our MADF. MADF learns to dynamically derive filters from mask (or feature maps of mask) for each of its convolution windows rather than relying on any predefined scaling factors and learns to update mask features automatically, so it is more flexible in handling the variously shaped valid pixels on each window. We will show the superiority of MADF over PConv in our experiment.
}

\begin{figure*}[!t]
    \centering
    \includegraphics[width=0.9\textwidth]{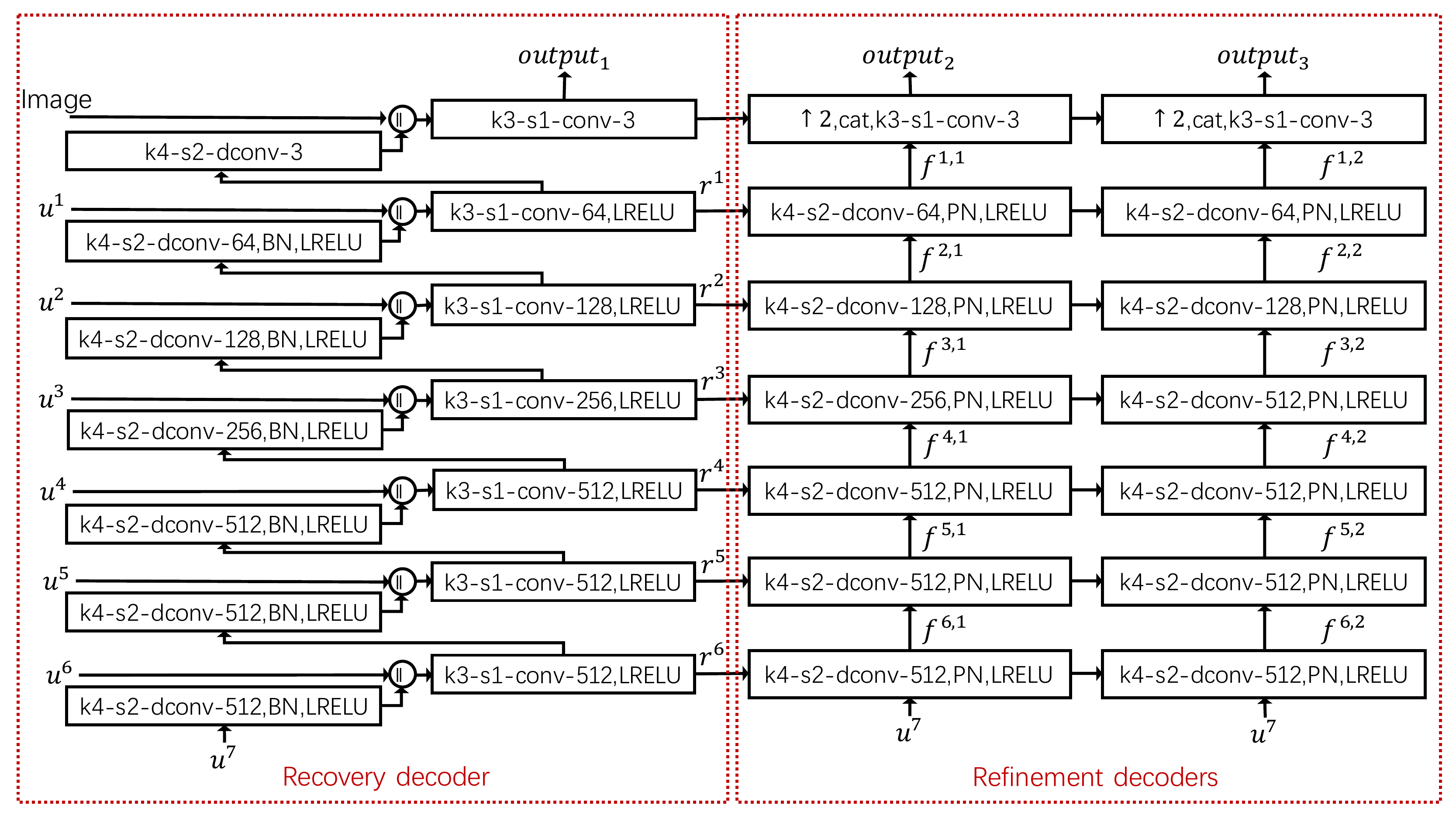}
    \caption{{\color{black}{The details of recovery decoder and refinement decoders. ``dconv-*'' denotes transposed convolutional layer with * output channels. ``PN'' is the point-wise normalization module and all convolution layers therein are with kernel size of 3 and stride of 1. Note that ``$\uparrow2$'' means $2\times$ upsample and is applied to $f^{l,k}$ in refinement decoders and ``cat'' means channel concatenation.}}}
    \label{fig:decoder}
\end{figure*}

\textbf{Difference from Gated Convolution}. 
\textcolor{black}{  
GConv is designed to generalize the hand-crafted renormalization strategy of PConv to a learning-based one. GConv performs "one set of filters for all windows" and relies on learned attention map for feature modulation. GConv uses shared convolutional kernels at different conv locations, then it assigns a scaling factor to each location. The kernels corresponding to different locations differ from each other with barely a scaling factor. However, the kernels of MADF for different convolution locations are generated dynamically according to mask, \textit{i.e.}, how many locations (windows) there are, how many sets of kernels there will be. 
{\color{black}{The design of MADF is more flexible than re-scaling vanilla convolution output of each convolution window and it is orthogonal to the gating mechanism proposed in GConv}.}
It is noteworthy that, MADF is computationally more efficient than GConv.
Learning attention maps for modulating features requires not only modules to generate attention maps but also element-wise multiplication per channel on per-point basis. We will provide quantitative analysis in Section~\ref{subsec:quan_compare_sota}.
{\color{black}{Besides, MADF is compatible with gating mechanism. Combining both of them may further improve the performance if extra computational cost of gating mechanism is not a major concern}.}
}

\subsection{Decoders}
\textit{Recovery Decoder.}
The recovery decoder aims to {\color{black}{roughing}} out multi-scale feature maps $r^1,\ldots, r^{L-1}$.
The $l^{th}$ recovery decoder block $R^l$ takes $r^l$ and $u^{l-1}$ as the inputs and generates $r^{l-1}$ as outputs. In detail, a transposed convolutional layer with stride 2 is used to up-sample $r^l$, which is then concatenated with $u^{l-1}$. Then a convolution is applied after the concatenation to further blend the $u^{l-1}$ and $r^l$ and leaky ReLU is used as the nonlinear activation, as shown in Figure \ref{fig:framework}(e). 
The U-shaped architecture allows us to fuse the information of $r^l$ and $u^{l-1}$ level by level.

\textit{Refinement Decoders.}
After the recovery decoder, we employ a sequence of refinement decoders $\{F_1, F_2,\ldots, F_K\}$ for further feature refinement. 
Each refinement decoder has $L-1$ refinement decoding blocks and generates $L-1$ feature maps of different scales. 
We use $F^{l,k}$ to indicate the $l^{th}$ block of $F^k$ and use $f^{l,k}$ to indicate the corresponding $l^{th}$ feature map. 

Because cascaded refinement framework is used and supervision information, such as reconstruction loss, perceptual loss, total variance loss etc., is incrementally leveraged to boost the inpainting results, we infer that statistical characteristics (mean value of $\mu$ and standard variance $\sigma$) of features of the hole regions and those of the non-hole regions are different, and so is the case for different refinement levels. Applying traditional batch normalization would neglect such difference and attenuate the effect of ``covariant shift removal''. To address this problem, we utilize Point-wise Normalization (PN) in the refinement phase to convolve the high-level feature $f^{l-1,k-1}$ from the prior decoder $F_{k-1}$ to dynamically produce the mask-aware scale and bias of batch normalization for the current decoder $F_k$. Mask awareness can be transmitted to the next refinement levels through PN.
Taking $F^{l,k}$ as an example, $f^{l,k}$ is firstly passed through transposed convolutional layer, and then the results $x$ are processed by a batch normalization layer whose affine transformation parameters $\alpha$ and $\beta$ are conditioned on $f^{l-1,k-1}$. 
Formally, PN behaves as the following equation shows:
\begin{equation}
     y = \alpha(f^{l-1,k-1})\cdot(x-\mu)/\sigma+\beta(f^{l-1,k-1}),
\end{equation}
where $\cdot$ means element-wise multiplication, {\color{black}{$x$ in the input to PN and $y$ is its output}}.
Specifically, $f^{l-1,k-1}$ is firstly projected onto an latent space, and then convolved to produce scale and bias tensors, which are of the same spatial dimension as $x$. The scale tensor is element-wise multiplied to the normalized activation of $f^{l,k}$ and the bias tensor is element-wise added to the result. Though mean and variance are calculated with both hole and non-hole regions, it can be adjusted via the adaptive point-wise scaling factor and bias term, achieving better normalization. 

{\color{black}{\subsection{Detailed Network Configurations}
We demonstrate detailed configurations of the encoder, recovery decoder and refinement decoders using block-diagrams in Figure \ref{fig:encoder} and Figure \ref{fig:decoder}, respectively. In these two figures, convolution kernel size, stride, number of output channels as well as activation function of each layer at every level is described in detail. The kernel sizes and strides are set empirically meanwhile channel numbers are carefully designed with model complexity kept in mind. Specifically, at the encoder, $C_m^l$ is set to a relatively small value of 16 and output channel number $C_e^l$ is limited by $\min(128, 16*2^l)$ for any $l$. Then we increase the channels of each $e^l$ by $1\times1$ convolution with ReLU activation to produce $u^l$. At each decoder, we set base channel number to be 64 and increase channel numbers by $2\times$ at every deeper level and the maximum channel number is also limited to be 512 to save computational cost. }}

\subsection{Loss Functions}
We train the network with the same four types of loss functions as partial convolution \cite{liu2018image}. However, the losses are added incrementally to the decoders. 

\textbf{Per-pixel Reconstruction Loss} Given the mask $M$, the result $I_{out}$, and the ground-truth image $I_{gt}$, we define the ${L}_1$ loss for the hole and outside-hole pixels as ${L}_{hole}$ and ${L}_{valid}$ respectively: 
\begin{equation}
L_{hole}=\frac{1}{N_{I_{gt}}}\|(1-M)\odot(I_{out}-I_{gt})\|_1
\end{equation}
\begin{equation}
L_{valid}=\frac{1}{N_{I_{gt}}}\|M\odot(I_{out}-I_{gt})\|_1
\end{equation}
\noindent where $N_{I_{gt}}$denotes the number of elements in the ground truth image $I_{gt}$ and {\color{black}{$M$ is the binary mask with zeros indicating missing pixels}}.

\textbf{Perceptual Loss} A VGG-based perceptual loss forces the decoder to generate image semantically closer to the ground-truth \cite{gatys2015neural}. We use layer \emph{pool}1, \emph{pool}2, \emph{pool}3 of the ImageNet pretrained VGG-16 \cite{simonyan2014very} for our loss calculation,
\begin{equation}
\begin{split}
L_{perc}=
&\sum_{p=0}^{P-1}\frac{\|\Psi_p^{I_{out}}-\Psi_p^{I_{gt}}\|_1}{N_{\Psi_p^{I_{gt}}}} \\
&+\sum_{p=0}^{P-1}\frac{\|\Psi_p^{I_{com}}-\Psi_p^{I_{gt}}\|_1}{N_{\Psi_p^{I_{gt}}}}
\label{equ:perc_loss}
\end{split}
\end{equation}
\noindent where $P$ denotes the number of layers selected from VGG, here it is 3. $I_{com}$ is the composition of the hole pixels from the raw output image $I_{out}$ and the non-hole pixels from the ground truth. $\Psi_p^{I_*}$ denotes the activation of the $p$-th selected VGG layer given the input $I_*$. ${N_{\Psi_p^{I_{gt}}}}$ is the dimension of the feature vector ${\Psi_p^{I_{gt}}}$ and is used as a normalization factor.

\textbf{Style Loss} A VGG-based style loss is sort of similar to the perceptual loss, but we perform an auto-correlation (Gram matrix) on each selected VGG feature map before applying $L_1$. Style loss makes the texture of generated image similar to that of the ground-truth. 
\begin{equation}
\begin{split}
L_{style}=
&\sum_{p=0}^{P-1}\frac{\|K_p((\Psi_p^{I_{out}})^\mathsf{T}(\Psi_p^{I_{out}})
-(\Psi_p^{I_{gt}})^\mathsf{T}(\Psi_p^{I_{gt}}))\|_1}{C_pC_p}\\
&+\sum_{p=0}^{P-1}\frac{\|K_p((\Psi_p^{I_{com}})^\mathsf{T}(\Psi_p^{I_{com}})
-(\Psi_p^{I_{gt}})^\mathsf{T}(\Psi_p^{I_{gt}}))\|_1}{C_pC_p}
\end{split}
\end{equation}
\noindent here, $(C_p, H_p, W_p)$ denotes the shape of the $\Psi_p^*$. $K_p$ equals to $1 / C_pH_pW_p$ for normalization.

\textbf{Total Variation Loss} It acts as the smoothing penalty \cite{johnson2016perceptual}. 
\begin{equation}
\begin{split}
L_{tv}=
&\sum_{(i,j)\in R,(i,j+1)\in R}\frac{\|I_{com}^{i,j+1}-I_{com}^{i,j}\|_1}{N_{I_{com}}}\\
&+\sum_{(i,j)\in R,(i+1,j)\in R}\frac{\|I_{com}^{i+1,j}-I_{com}^{i,j}\|_1}{N_{I_{com}}}
\end{split}
\end{equation}
\noindent where $R$ represents the 1-pixel dilated hole regions.

The above losses are incrementally leveraged to train our cascaded refinement framework. Specifically, for the output of the recovery decoder, we use $L_1$ loss:
\begin{equation}
L_1=L_{valid}+6L_{hole}.
\label{equ:l1_loss}
\end{equation}
For the output of the first refinement decoder, we use L1 loss and perceptual loss: $L_1+0.05L_{perc}$, and the output of last refinement decoder is supervised by $L_{total}$:
\begin{equation}
L_{total}=L_1+0.05L_{perc}+120L_{style}+0.1L_{tv}
\label{equ:total_loss}
\end{equation}
The loss weights are empirically set by following the practice in \cite{liu2018image}.

\section{Experiments}
\begin{table*}[!ht]
    \caption{Results over Places2 dataset, ``Regular" means a $128\times128$ rectangle mask in the center of image. ``ALL" indicates measuring in the whole testset. $^\dag$ means higher is better, $^\star$ means lower is better. }
    \label{tab:comparison_place2}
    \centering
    \begin{tabular}{l|l|ccccccc|c} \hline
         & Methods & (0.01,0.1] & (0.1,0.2]  & (0.2,0.3] & (0.3,0.4]  & (0.4,0.5] & (0.5,0.6]  & ALL & Regular \\ \hline
         \hline
          & PM {\color{black}{\cite{barnes2009patchmatch}}}  & 32.28 & 26.37 & 23.55 & 21.40 & 19.70 & 17.58 & 23.48 & 19.70\\
          & CA {\color{black}{\cite{yu2018generative}}} & 29.67 & 23.81 & 20.74 & 18.68 & 17.32 &15.94 &21.03 & 19.52\\
          & EC {\color{black}{\cite{nazeri2019edgeconnect}}} & 33.58	&27.87	&24.89	&22.79	&21.10	&18.95 & 24.86 &20.44\\
$PSNR^\dag$ &  PConv {\color{black}{\cite{liu2018image}}} &34.05&28.42&	25.54&	23.24&	21.42&	19.10& 25.30&21.06\\
          & GConv {\color{black}{\cite{yu2019free}}} & 32.45 & 26.68 & 23.73 & 21.45 & 19.70 & 17.40 & 23.57 & 19.11\\
          &  LBAM {\color{black}{\cite{chaohaoLBAM2019}}} & 34.09&	28.13&	25.13&	22.83&	21.02&	18.77& 25.00&20.68\\
          & {\color{black}{PEN-Net \cite{pennet}}} &{\color{black}{31.61}}& {\color{black}{25.76}} & {\color{black}{23.04}} & {\color{black}{21.07}} & {\color{black}{19.39}} & {\color{black}{18.29}} & {\color{black}{23.19}} & {\color{black}{20.20}} \\
          & {\color{black}{StructureFlow \cite{structureflow}}}&{\color{black}{34.92}} &{\color{black}{29.13}} &{\color{black}{25.89}} &{\color{black}{23.58}} &{\color{black}{21.63}} &{\color{black}{19.35}} &{\color{black}{25.77}} &{\color{black}{21.23}} \\
           &  Ours  & \textbf{35.74}&\textbf{ 29.42}&\textbf{26.29}&\textbf{23.84}&	\textbf{21.92}&	\textbf{19.44}& \textbf{26.11}&\textbf{21.32} \\
\hline
         &  PM {\color{black}{\cite{barnes2009patchmatch}}}& 0.979&	0.935&	0.877&	0.806&	0.724&	0.573&	0.816&0.717 \\
         &  CA {\color{black}{\cite{barnes2009patchmatch}}}& 0.961&	0.893&	0.804&	0.715&	0.628&	0.511&	0.752& 0.735\\
         &  EC {\color{black}{\cite{nazeri2019edgeconnect}}} & 0.980&	0.945&	0.895&	0.837&	0.768&	0.634&	0.843&0.747\\
$SSIM^\dag$ & PConv {\color{black}{\cite{liu2018image}}} &0.983&	0.951&	0.908&	0.853&	0.788&	0.652&	0.856&0.756  \\
          & GConv {\color{black}{\cite{yu2019free}}} & 0.979 & 0.939 & 0.886 & 0.822 & 0.751 & 0.609 & 0.831 & 0.726\\
          &  LBAM {\color{black}{\cite{chaohaoLBAM2019}}} & 0.983&	0.949&	0.903&	0.845&	0.777&	0.638&	0.849&0.748 \\
          & {\color{black}{PEN-Net \cite{pennet}}} &{\color{black}{0.976}} &{\color{black}{0.910}} &{\color{black}{0.835}} &{\color{black}{0.753}} &{\color{black}{0.6456}} &{\color{black}{0.547}} &{\color{black}{0.778}} & {\color{black}{0.726}}  \\
          & {\color{black}{StructureFlow \cite{structureflow}}}&{\color{black}{0.985}} &{\color{black}{0.957}} &{\color{black}{0.913}} &{\color{black}{0.861}} &{\color{black}{0.796}} &{\color{black}{0.654}} &{\color{black}{0.863}} &{\color{black}{0.767}} \\
          &  Ours  &\textbf{0.988}&\textbf{0.961}&\textbf{0.922}&\textbf{0.873}&	\textbf{0.812}&\textbf{0.679}&\textbf{0.872}&\textbf{0.768}\\
    \hline
          &  PM {\color{black}{\cite{barnes2009patchmatch}}} & 4.99&12.69&22.33&	31.94&43.42&52.29&	11.38&12.68 \\
          &  CA {\color{black}{\cite{barnes2009patchmatch}}}& 7.13&16.25&28.48&	40.31&52.89&60.46&	15.84&11.42\\
          &  EC {\color{black}{\cite{nazeri2019edgeconnect}}} & 4.06&8.78&14.51&20.25&27.30&37.52&	5.76&10.87\\
$FID ^\star$ & PConv {\color{black}{\cite{liu2018image}}} & 4.64&10.01&16.47&22.93&31.69&	43.31&7.53&15.91\\
          & GConv {\color{black}{\cite{yu2019free}}} & 4.18 & 9.93& 16.93& 24.25& 32.44& 44.42& 7.88 & 13.22\\
          &  LBAM {\color{black}{\cite{chaohaoLBAM2019}}}& 4.21&9.34&15.69&22.25&30.31&42.38&	7.06&12.07 \\
          & {\color{black}{PEN-Net \cite{pennet}}} &{\color{black}{4.32}} &{\color{black}{11.81}} &{\color{black}{19.16}} &{\color{black}{27.27}} &{\color{black}{35.44}} &{\color{black}{48.73}} &{\color{black}{8.29}} &{\color{black}{10.91}} \\
          & {\color{black}{StructureFlow \cite{structureflow}}}&{\color{black}{3.11}} &{\color{black}{6.82}} &{\color{black}{13.78}} &{\color{black}{19.82}} &{\color{black}{24.97}} &{\color{black}{37.33}} &{\color{black}{5.54}} &\textbf{{\color{black}{8.24}}} \\
          &  Ours & \textbf{2.69}&\textbf{6.71}&\textbf{11.64}&\textbf{16.95}&\textbf{23.27}&\textbf{34.13}&\textbf{4.45}&{8.29}\\
\hline
    \end{tabular}

\end{table*}
In this section, we first introduce the experimental settings, then we conduct ablation study to verify the effectiveness of design choices in our architecture. Finally, we compare with other methods from objective quantitative and subjective perception aspects to show strength of our method.
\subsection{Datasets}
We evaluate our model on three well-known public datasets. 
\begin{itemize}
\item Places2 \cite{zhou2017places}, a collection contains about millions images comprising 365 unique scene categories.  
We use the high-resolution version images, including 1.8 million training images and 12K testing images.
\item CelebA \cite{liu2015deep}, 
a face attributes dataset containing about 200K images. 
We use 200K images for training, and the remaining images for testing.
\item {\color{black}{Paris StreetView}} \cite{doersch2015makes}, a dataset of highly-structure facades collected from the Google Street View in Paris.
We use 14K images for training, and 100 images for testing. 
\end{itemize}

\subsection{Reference State-of-the-Arts}
We compare our framework with {\color{black}{8}} recent state-of-the-art ones. For the purpose of fairness, we try to directly use the officially released pre-trained model with the same experimental settings as much as possible. For PatchMatch \cite{barnes2009patchmatch} (PM), we use a third-party implementation. For Contextual Attention\cite{yu2018generative} (CA), EdgeConnect \cite{nazeri2019edgeconnect} (EC) and Gated convolution \cite{yu2019free} (GConv), we directly use the officially release pre-trained model. Since the source code of Partial convolution \cite{liu2018image} (Pconv) is not available, we implement it with the experimental settings in the paper. Bidirectional Attentional Maps \cite{chaohaoLBAM2019} (LBAM) is tested with the source code and the pertrained models provided by its authors. {\color{black}{Performances of PEN-Net \cite{pennet} and StructureFlow \cite{structureflow} are also evaluated using their official released codes and models.}}

\subsection{Implementation details and training process}
We use the same training and testing irregular masks as  \cite{liu2018image}, including 55,116 training masks and 12,000 testing masks. 
The testing masks can be categorized into 6 categories through hole-to-image area ratios: (0.01,0.1], (0.1,0.2], (0.2,0.3], (0.3,0.4], (0.4,0.5], (0.5,0.6].
During training we augment the mask set by randomly flipping, cropping, rotating and dilating.
We train the model by the Adam\cite{kingma2014adam} optimizer with $\beta_1=0.9$ and $\beta_2=0.999$ and learning rate is set to 0.0002. All the models are trained on 8 NVIDIA V100 GPUs (32G) with batch size of 24. We train our model for about 300K iterations. The training process takes about 5 days. 
{\color{black}{The weight decay factor is set to be 0 by default and the learning rate policy is a fixed one, \textit{i.e.}, learning rate is a constant for each of the 300K iterations. There is no appropriate pretrained model for initialization. We train our model from scratch and its model parameters are randomly initialized from a normal distribution, the mean and variance of the normal distribution is 0 and 0.01, respectively.}
}
Our model does not require any post-processing.

\subsection{Ablation study}
\begin{figure*}[t]

\begin{center}
\setlength{\tabcolsep}{0.5mm}{
\begin{tabular}{ccccccccccccccccc}
\centering
\includegraphics[width=0.05\textwidth]{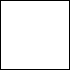}&
\includegraphics[width=0.05\textwidth]{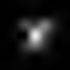}&\includegraphics[width=0.05\textwidth]{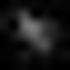}&\includegraphics[width=0.05\textwidth]{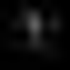}&\includegraphics[width=0.05\textwidth]{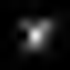}&\includegraphics[width=0.05\textwidth]{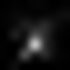}&\includegraphics[width=0.05\textwidth]{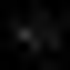}&\includegraphics[width=0.05\textwidth]{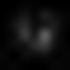}&\includegraphics[width=0.05\textwidth]{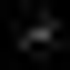}&\includegraphics[width=0.05\textwidth]{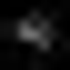}&\includegraphics[width=0.05\textwidth]{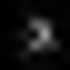}&\includegraphics[width=0.05\textwidth]{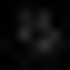}&\includegraphics[width=0.05\textwidth]{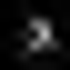}&\includegraphics[width=0.05\textwidth]{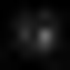}&\includegraphics[width=0.05\textwidth]{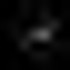}&\includegraphics[width=0.05\textwidth]{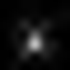}&\includegraphics[width=0.05\textwidth]{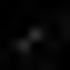}\\
\includegraphics[width=0.05\textwidth]{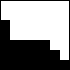}&
\includegraphics[width=0.05\textwidth]{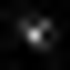}&\includegraphics[width=0.05\textwidth]{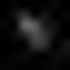}&\includegraphics[width=0.05\textwidth]{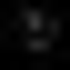}&\includegraphics[width=0.05\textwidth]{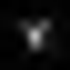}&\includegraphics[width=0.05\textwidth]{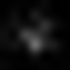}&\includegraphics[width=0.05\textwidth]{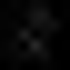}&\includegraphics[width=0.05\textwidth]{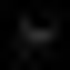}&\includegraphics[width=0.05\textwidth]{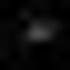}&\includegraphics[width=0.05\textwidth]{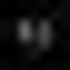}&\includegraphics[width=0.05\textwidth]{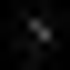}&\includegraphics[width=0.05\textwidth]{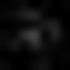}&\includegraphics[width=0.05\textwidth]{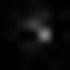}&\includegraphics[width=0.05\textwidth]{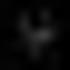}&\includegraphics[width=0.05\textwidth]{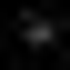}&\includegraphics[width=0.05\textwidth]{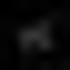}&\includegraphics[width=0.05\textwidth]{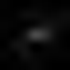}\\
\includegraphics[width=0.05\textwidth]{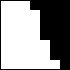}&
\includegraphics[width=0.05\textwidth]{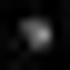}&\includegraphics[width=0.05\textwidth]{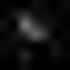}&\includegraphics[width=0.05\textwidth]{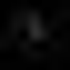}&\includegraphics[width=0.05\textwidth]{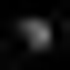}&\includegraphics[width=0.05\textwidth]{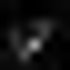}&\includegraphics[width=0.05\textwidth]{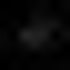}&\includegraphics[width=0.05\textwidth]{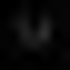}&\includegraphics[width=0.05\textwidth]{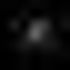}&\includegraphics[width=0.05\textwidth]{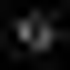}&\includegraphics[width=0.05\textwidth]{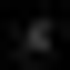}&\includegraphics[width=0.05\textwidth]{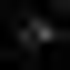}&\includegraphics[width=0.05\textwidth]{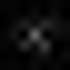}&\includegraphics[width=0.05\textwidth]{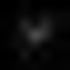}&\includegraphics[width=0.05\textwidth]{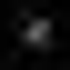}&\includegraphics[width=0.05\textwidth]{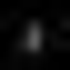}&\includegraphics[width=0.05\textwidth]{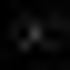}\\
\includegraphics[width=0.05\textwidth]{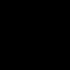}&
\includegraphics[width=0.05\textwidth]{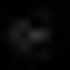}&\includegraphics[width=0.05\textwidth]{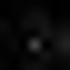}&\includegraphics[width=0.05\textwidth]{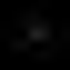}&\includegraphics[width=0.05\textwidth]{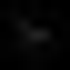}&\includegraphics[width=0.05\textwidth]{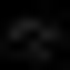}&\includegraphics[width=0.05\textwidth]{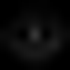}&\includegraphics[width=0.05\textwidth]{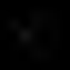}&\includegraphics[width=0.05\textwidth]{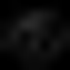}&\includegraphics[width=0.05\textwidth]{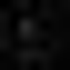}&\includegraphics[width=0.05\textwidth]{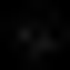}&\includegraphics[width=0.05\textwidth]{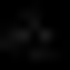}&\includegraphics[width=0.05\textwidth]{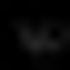}&\includegraphics[width=0.05\textwidth]{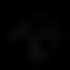}&\includegraphics[width=0.05\textwidth]{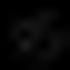}&\includegraphics[width=0.05\textwidth]{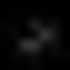}&\includegraphics[width=0.05\textwidth]{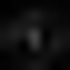}\\
\small{a}&\small{b}&\small{c}&\small{d}&\small{e}&\small{f}&\small{g}&\small{h}&\small{i}&\small{j}&\small{k}&\small{l}&\small{m}&\small{n}&\small{o}&\small{p}&\small{q}\\

\end{tabular}}
\end{center}
\caption{Visualization of the generated 16 kernels of the 1st conv layer in the encoder. The corresponding mask patch is shown in the first column, and the missing regions are shown in black. The original kernels are $7\times7$ and are resized to $70\times 70$.}
\label{fig:visual_kernel}
\end{figure*}
We implement several variants of our framework and conduct ablation experiments on Places2 testset. We use 12K testing images with random masks. As previous image inpainting works do, we measure our models in terms of PSNR, SSIM\cite{wang2004image} and Frechet Inception Distance (FID) \cite{heusel2017gans}. PSNR and SSIM are pixel level evaluation metrics while FID measures the Wasserstein-2 distance between real and fake images. 
\subsubsection{Design Choice of Network Architecture}
We carry out experiments to analyze effectiveness of each component of our network, including MADF, PN and cascaded refinement decoder. Both qualitative and quantitative results (see Table \ref{tb:num_of_refine_d}) are provided for comparison. In Tabel \ref{tb:num_of_refine_d}, ``PConv'' denotes partial convolution.

\textbf{MADF v.s. PConv}
From Table \ref{tb:num_of_refine_d},  it can be seen that using the basic U-shaped encoder-decoder framework (\emph{i.e.}, the number of refinement decoder is 0), MADF outperforms partial convolution.
The PSNR is improved from 25.49dB to 25.70dB by replacing partial convolution with our proposed mask-aware dynamic filtering. 

\begin{table}[t]
\caption{Ablation study on model components. $^\dag$ means higher is better, $^\star$ means lower is better. $K*$ denotes the number of the refinement decoders. ``PN'' denotes using point-wise normalization. 
}
\label{tb:num_of_refine_d}
\centering
\begin{tabular}{lcccc} \hline
 & $PSNR^\dag$  & $SSIM^\dag$ & $FID^\star$ 	\\ \hline
\hline
PConv+K0 & 25.49 & 0.860 & 5.82 \\
GConv+K0 & 25.61 & 0.863 & 5.53 \\
MADF+K0 & 25.70 & 0.864 &  5.21 \\
MADF+K1 PN& 25.87 & 0.867 & 5.03 \\
MADF+K1 & 25.75 & 0.865 & 5.23 \\
MADF+K2 PN& 26.11 & 0.872 & \textbf{4.45} \\
MADF+K2 & 25.96 & 0.869 & 4.96 \\
MADF+K3 PN& \textbf{26.16} & \textbf{0.874} & 4.69 \\
\hline
\end{tabular}
\end{table}

\begin{table}[t]
\caption{Impact of supervision schemes for intermediate decoder. $^\dag$ means higher is better, $^\star$ means lower is better.
}
\label{tb:losses_effect}
\centering
\begin{tabular}{ccccc} \hline
Loss & $PSNR^\dag$  & $SSIM^\dag$ & $FID^\star$ 	\\ \hline
\hline 
None   & 25.97 & 0.868  & 4.99 \\
Same 	& 26.02 & 0.869 & 4.84 \\
Coarse-to-fine   & \textbf{26.11} & \textbf{0.872} & \textbf{4.45} \\
\hline
\end{tabular}

\end{table}
\textcolor{black}{  
Furthermore, we visualize 16 $7\times 7$ MADF kernels generated from their corresponding mask regions at the first convolution layer of the encoder in Figure~\ref{fig:visual_kernel}. It does show that our model has learned to dynamically generate appropriate kernels according to the mask. Specifically, each row in Figure~\ref{fig:visual_kernel} represents a region of the input and each column corresponds to one of the 16 kernels. For example, the first row corresponds to a region that all pixels are valid and the energy of most kernels is high. While, the last row corresponds to a region where all pixels are masked, the energy of all the kernels is very small. In the second and third rows, where partial pixels are missing, if we check column (b) (e) (f) and (m), we can see that the energy at the valid part is higher than the energy at the masked part.  
}

\textcolor{black}{   
\textbf{MADF v.s. GConv}
In Table 1, MADF+K0 is an UNet architecture without refinement decoders, which achieves average PSNR of 25.7dB. 
We get a variant by replacing the MADF module in MADF+K0 with GConv, which is denoted as GConv+K0. Following the same training strategies and experimental settings, GConv+K0 achieves average PSNR of 25.61dB, SSIM of 0.863 and FID of 5.53, which is inferior to MADF+K0. This validates the effectiveless of MADF.
}

\textbf{PN v.s. BN}
The results in the third row and fifth row demonstrate the performance with our proposed PN module. Compared to the BN counterpart, the PSNR and SSIM index are significantly improved, showing that point-wise normalization is able to capture the discrepancy among different refinement levels, and learn the adaptive scale and bias parameters for assisting batch normalization.
Besides, PN constantly outperforms BN whatever the number of refinement decoders is.

\textbf{Cascaded Refinement}
Here we also show that empirically use two refinement decoders is a good design choice. From Table \ref{tb:num_of_refine_d}, we can see no matter whether PN is enabled or not, performances of our model will be consistently improved when adding refinement decoder until 3 of them are used. Meanwhile the 3rd refinement decoder gains few. So we finally use 2 refinement decoders.

\begin{figure}[h]
\begin{center}
\setlength{\tabcolsep}{0.5mm}
\begin{tabular}{cccc}
\centering
\includegraphics[width=0.23\columnwidth]{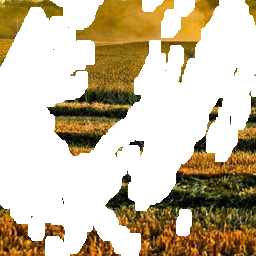}&\includegraphics[width=0.23\columnwidth]{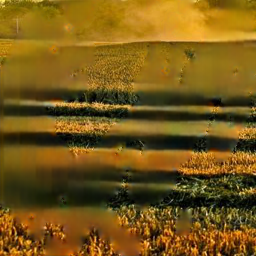}&\includegraphics[width=0.23\columnwidth]{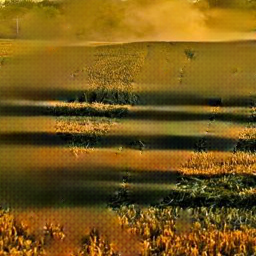}&\includegraphics[width=0.23\columnwidth]{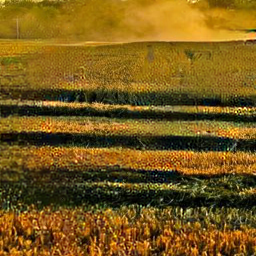} \\
\small{(a) Masked}& \small{(b) output of $R$}& \small{(c) output of $F^1$}& \small{(d) Output of $F^2$}\\
\end{tabular}
\end{center}
\caption{The ``coarse-to-fine'' supervision schema helps our framework to recover image.  Best viewed with zoom-in.}
\label{fig:coarse-to-fine-results}
\end{figure}
\subsubsection{Incremental Supervision} Each intermediate decoder produces an intermediate recovered result. Now, we explore whether the addition of intermediate supervisions will improve the performance.
We take intermediate results without supervisions as baseline, and compare with the ``coarse to fine" supervision schema and the ``same" supervision schema. The ``coarse to fine" schema denotes our strategy. For the ``same" schema, all the decoders take the same loss function of Eq.\ref{equ:total_loss}. It can be observed from Table \ref{tb:losses_effect}, ``same" schema contributes little to the performance, while ``coarse-to-fine" schema can bring obvious improvements. By dividing the challenging task into many easier sub-tasks, ``coarse-to-fine" schema helps the model to generate features and intermediate results in a coarse-to-fine way, as shown in Figure \ref{fig:coarse-to-fine-results}.

\begin{figure*}[!h]
\begin{center}
\setlength{\tabcolsep}{0.5mm}
\begin{tabular}{ccccccc}
\centering
\includegraphics[width=0.12\textwidth]{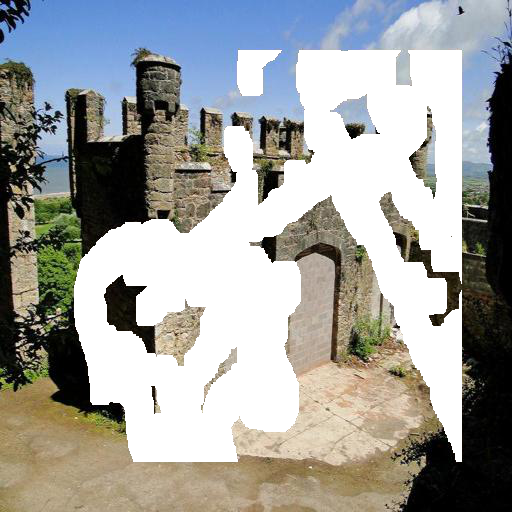}&
\includegraphics[width=0.12\textwidth]{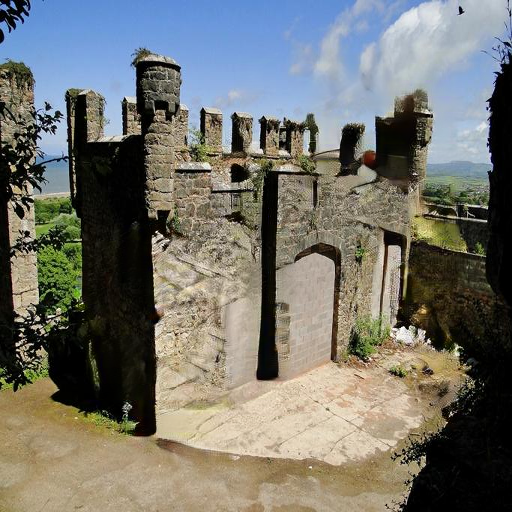}&
\includegraphics[width=0.12\textwidth]{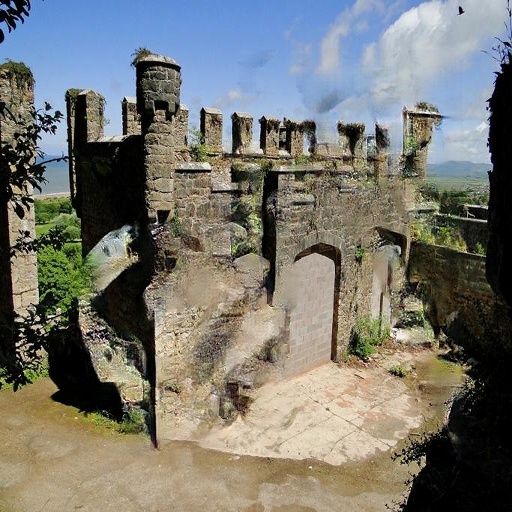}&
\includegraphics[width=0.12\textwidth]{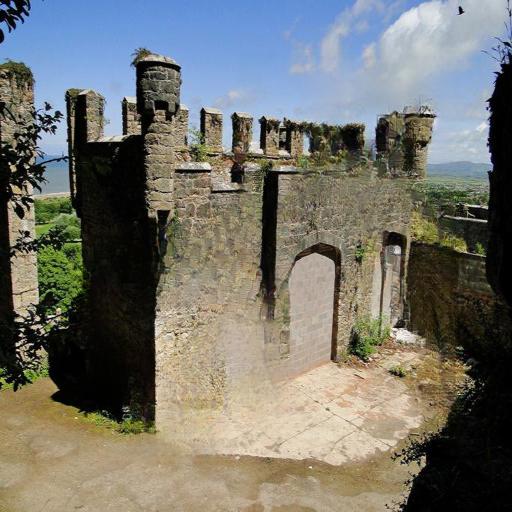}&
\includegraphics[width=0.12\textwidth]{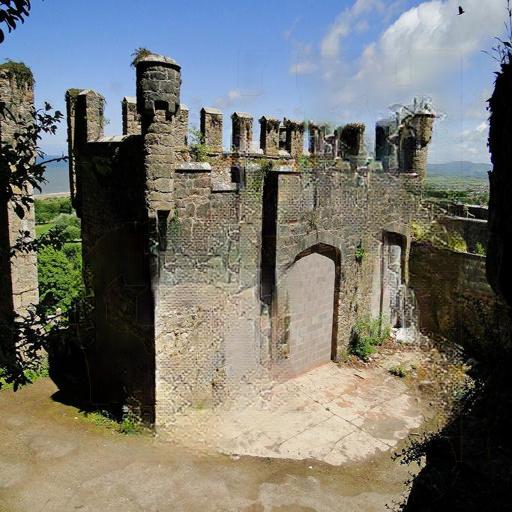}&
\includegraphics[width=0.12\textwidth]{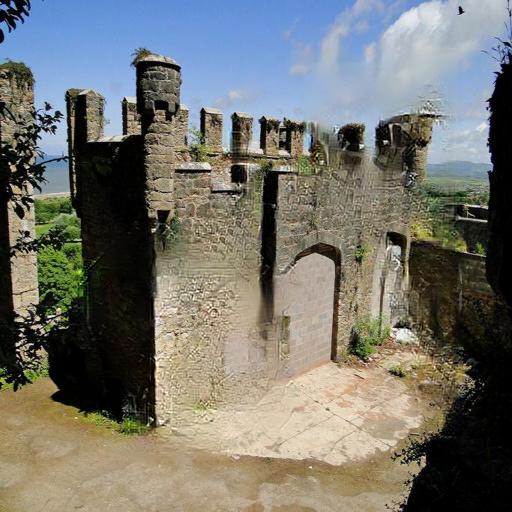}&
\includegraphics[width=0.12\textwidth]{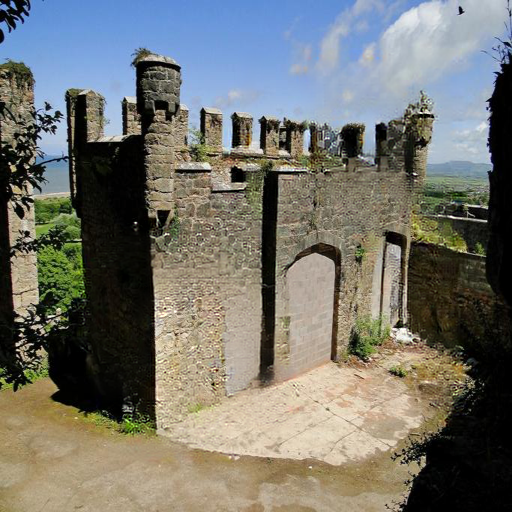}\\

\includegraphics[width=0.12\textwidth]{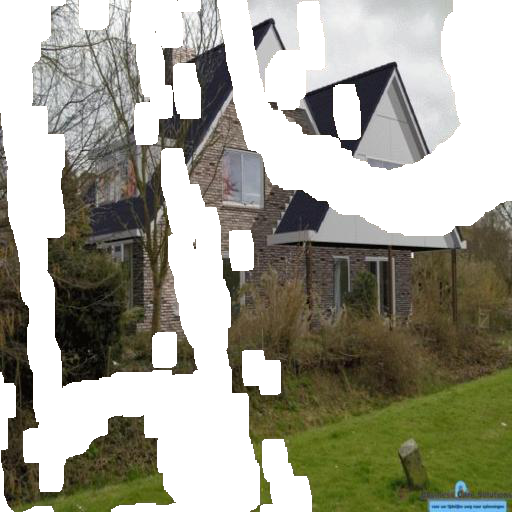}&
\includegraphics[width=0.12\textwidth]{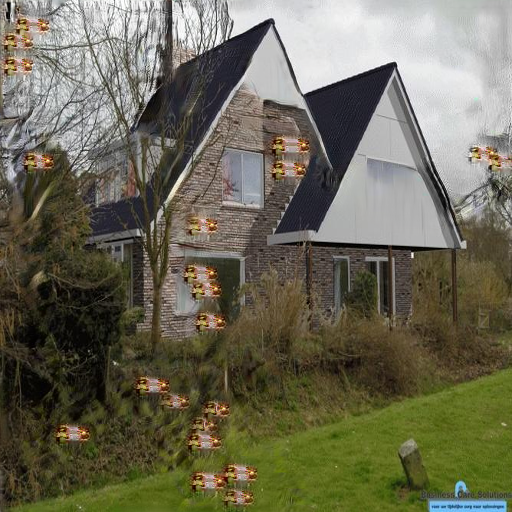}&
\includegraphics[width=0.12\textwidth]{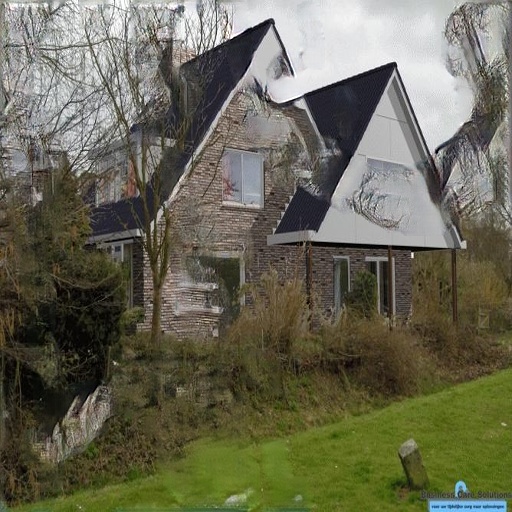}&
\includegraphics[width=0.12\textwidth]{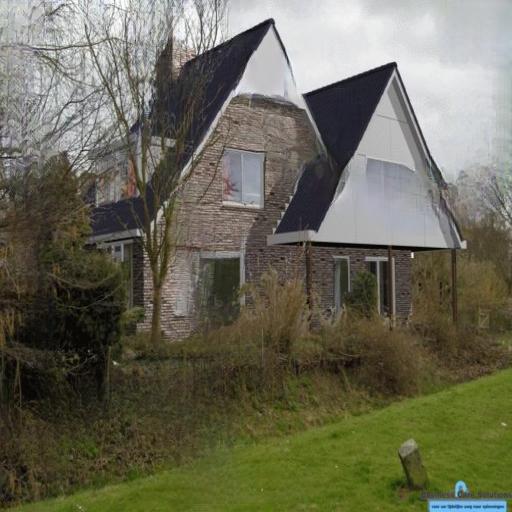}&
\includegraphics[width=0.12\textwidth]{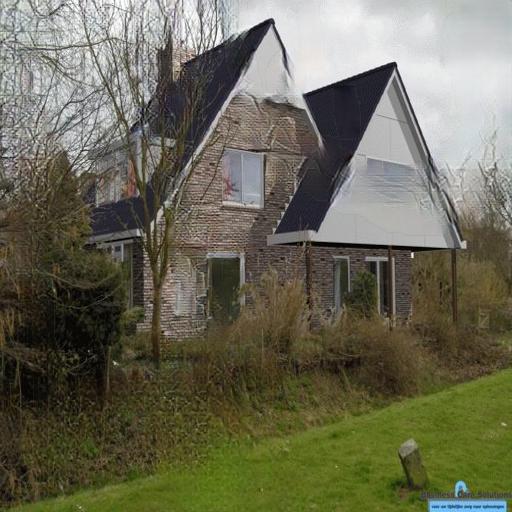}&
\includegraphics[width=0.12\textwidth]{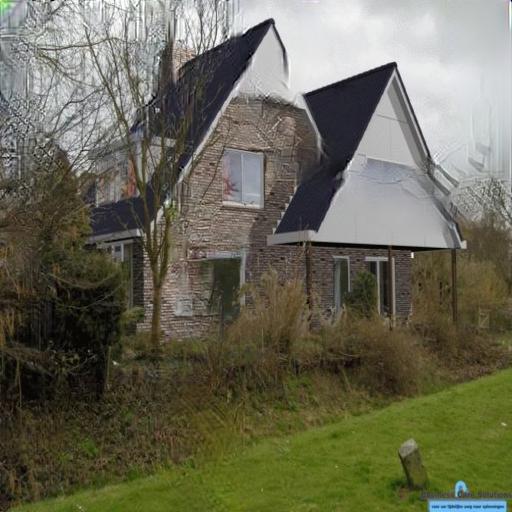}&
\includegraphics[width=0.12\textwidth]{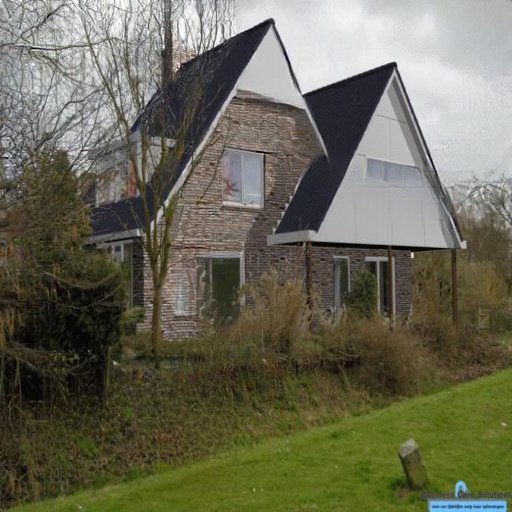}\\

\small{Input}& \small{GConv}& \small{CA}& \small{EC}&\small{LBAM}&\small{PConv}&\small{Ours}\\
\end{tabular}
\end{center}
\caption{Qualitative comparison on Places2. Best viewed with zoom-in.}
\label{fig:comp_on_places2}
\end{figure*}

\begin{figure*}[!h]
\begin{center}
\setlength{\tabcolsep}{0.5mm}
\begin{tabular}{ccccccc}
\centering
\includegraphics[width=0.12\textwidth]{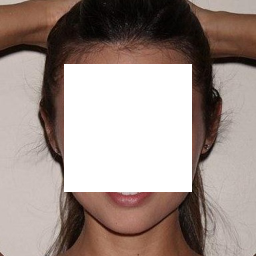}&
\includegraphics[width=0.12\textwidth]{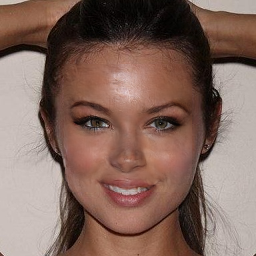}&
\includegraphics[width=0.12\textwidth]{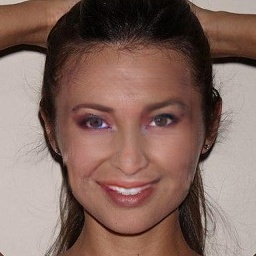}&
\includegraphics[width=0.12\textwidth]{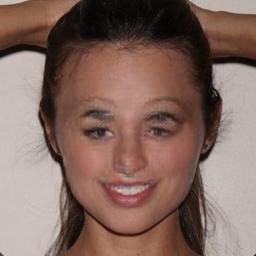}&
\includegraphics[width=0.12\textwidth]{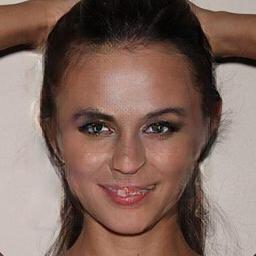}&
\includegraphics[width=0.12\textwidth]{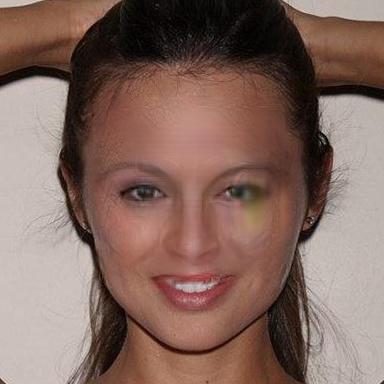}&
\includegraphics[width=0.12\textwidth]{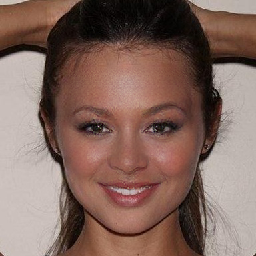}\\
\includegraphics[width=0.12\textwidth]{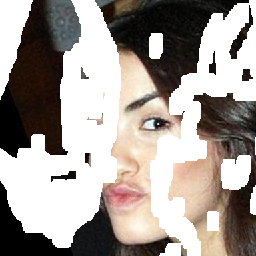}&
\includegraphics[width=0.12\textwidth]{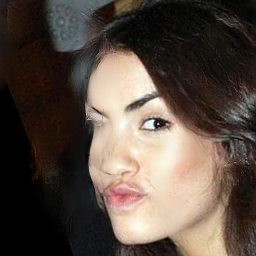}&
\includegraphics[width=0.12\textwidth]{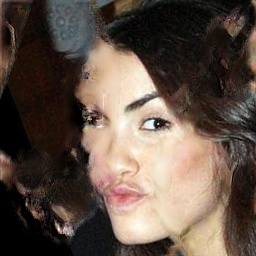}&
\includegraphics[width=0.12\textwidth]{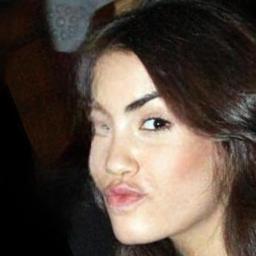}&
\includegraphics[width=0.12\textwidth]{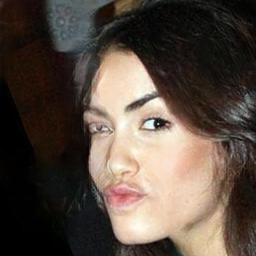}&
\includegraphics[width=0.12\textwidth]{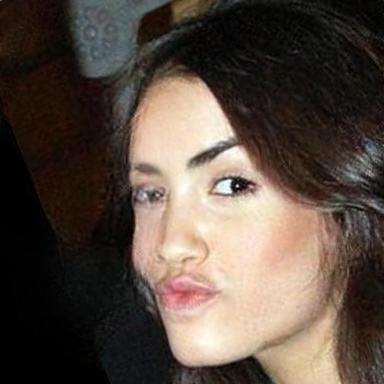}&
\includegraphics[width=0.12\textwidth]{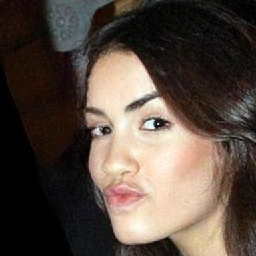}\\

\small{Input}& \small{GConv}& \small{CA}& \small{EC}&\small{LBAM}&\small{PConv}&\small{Ours}\\
\end{tabular}
\end{center}
\caption{Qualitative comparison on CelebA. Best viewed with zoom-in.}
\label{fig:comp_on_celeba}
\end{figure*}

\begin{figure*}[!h]
\begin{center}
\setlength{\tabcolsep}{0.5mm}
\begin{tabular}{ccccccc}
\centering
\includegraphics[width=0.12\textwidth]{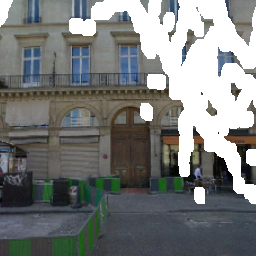}&
\includegraphics[width=0.12\textwidth]{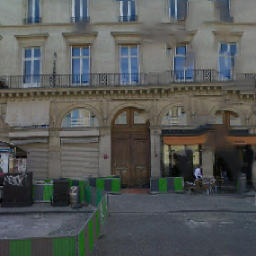}&
\includegraphics[width=0.12\textwidth]{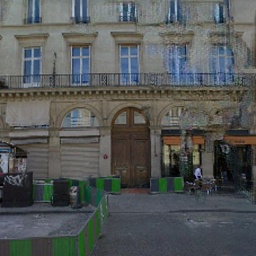}&
\includegraphics[width=0.12\textwidth]{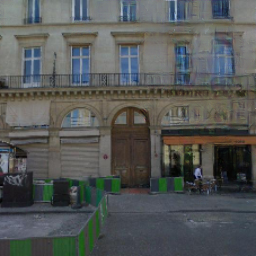}&
\includegraphics[width=0.12\textwidth]{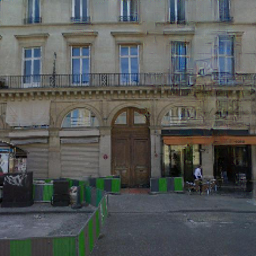}&
\includegraphics[width=0.12\textwidth]{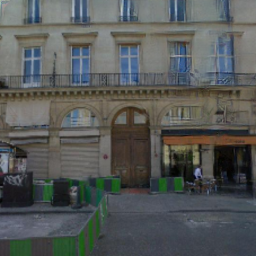}&
\includegraphics[width=0.12\textwidth]{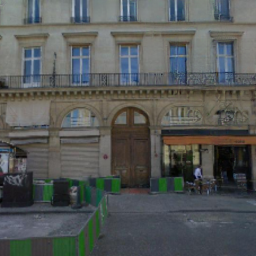}\\

\includegraphics[width=0.12\textwidth]{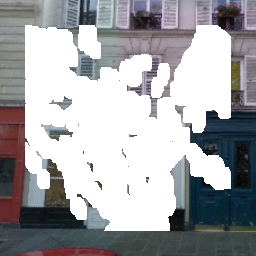}&
\includegraphics[width=0.12\textwidth]{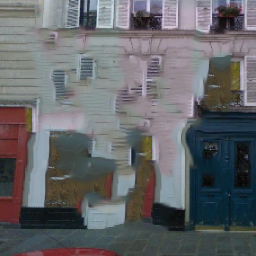}&
\includegraphics[width=0.12\textwidth]{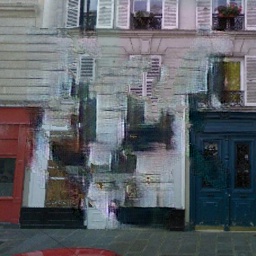}&
\includegraphics[width=0.12\textwidth]{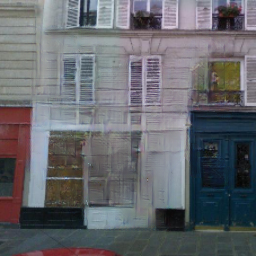}&
\includegraphics[width=0.12\textwidth]{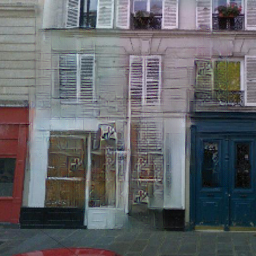}&
\includegraphics[width=0.12\textwidth]{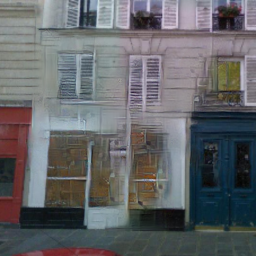}&
\includegraphics[width=0.12\textwidth]{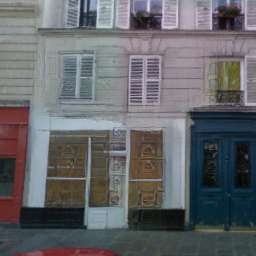}\\

\small{Input}& \small{PM}& \small{CA}& \small{EC}&\small{LBAM}&\small{PConv}&\small{Ours}\\
\end{tabular}
\end{center}
\caption{Qualitative comparison on {\color{black}{Paris StreetView}}. Best viewed with zoom-in.}
\label{fig:comp_on_psv}
\end{figure*}

\subsection{Quantitative Comparisons with SOTAs}
\label{subsec:quan_compare_sota} 
Places2 dataset contains various scenes, which is similar to the distribution of natural scene. 
So we compare with existing methods on this dataset for quantitative results. We use 12,000 testing images, which are randomly masked with the testing masks. Comparison experiments are conducted with irregular masks and regular masks respectively. Due to the measurements are sensitive to the size of image, we test all the methods with the same image size $256\times256$ following the common practice \cite{nazeri2019edgeconnect,yu2018generative,chaohaoLBAM2019}.  
Table \ref{tab:comparison_place2} shows the evaluation results. It can be seen that across different mask shapes and different missing areas, our model outperforms all the other state-of-the-art methods on these measurements.

{\color{black}{PSNR measures the L2 distance, SSIM measures structure similarity meanwhile FID is recently proposed to mimic human perception of similarity in images. It is worthy of noting that the PSNR, SSIM and FID achieved by our model are consistently superior. Besides, when small missing regions appear in the input image, the performance of our model is extremely good. For example, when missing ratio is lower than 0.1, the inpainting results demonstrate average PSNR of 35.74dB, SSIM of 0.988 and FID of 2.69. It shows our model can well capture the context information to fill in the small missing regions. When moderate regions are to be inpainted, the results are also plausible, for instance, when missing ratio is (0.2,0.3], the SSIM is 0.922. When a large portion of image is missing, say missing ratio is (0.5,0.6], the visual quality will be largely degraded and the FID metric is 34.13. This is also reasonable because the valid pixels are very limited in this case and context information is not sufficient for the model to fill in holes. Last but not least, missing ratio of regular hole equals 0.25, however, the performance on regular hole is much worse than free-form missing region with ratio of (0.2, 0.3]. It suggests that random missing region is easier to be filled than regular missing region.}}

\begin{table}[!t]
    \centering
    {\color{black}\begin{tabular}{c|c|c|c} 
    \hline
         Method &FLOPs &PSNR &FID  \\
         \hline
         PConv \cite{liu2018image}+K0       &\textbf{18.95G}    &25.49dB &5.82 \\
         LBAM \cite{chaohaoLBAM2019}       &22.11G    &25.00dB    &7.06 \\
         \textbf{MADF+K0}     &{22.13G}    &\textbf{25.70dB} &\textbf{5.21}\\
         \hline
         GConv \cite{yu2019free}       &\textbf{55.51G}    &23.57dB &7.88     \\
         EC \cite{nazeri2019edgeconnect}&122.67G   &24.86dB &5.76       \\
         CA \cite{yu2018generative}          &51.62G    &21.03dB &15.84       \\
         \textbf{MADF+K2+PN}  &{51.77G}    &\textbf{26.11dB} &\textbf{4.45}\\
         \hline
    \end{tabular}}
    \caption{Comparison on complexity-effectiveness trade-off among different methods.}
    \label{tab:my_label}
\end{table}
For comprehensive judging our work, we compare FLOPs of our  models  with  several  existing SOTAs {\color{black}{and the results are listed in Table \ref{tab:my_label}}}. {\color{black}{In this paper, we measure the model complexity by the well-known metric of ``FLOPs''. It measures the float point operations of a model. We report the number of float point multiplications instead of number of model parameters in this paper, because float point multiplication contributes much more to the model complexity compared to addition operation and it is a better indicator of model inference speed compared to number of model parameters.}}
Without  refinements  (MADF+K0  in  Table \ref{tb:num_of_refine_d}),  our  model  requires 22.13G FLOPs, which is slightly larger than previous SOTA PConv (18.95G), however, the performance is much better (25.70dB v.s.  25.49dB). Our final model (MADF+K2+PN in Table \ref{tb:num_of_refine_d})  requires  51.77G  FLOPs  and  performance  is boosted  to  26.11dB,  GConv,  LBAM,  EdgeConnect  and Contextual Attention requires 55.51G, 22.11G, 122.67G and 51.62G, respectively. Here FLOPs are calculated under test image resolution of $256\times256$.

\subsection{Qualitative Comparisons with SOTAs}

\textcolor{black}{    
Figure~\ref{fig:comp_on_places2}, Figure~\ref{fig:comp_on_celeba} and Figure~\ref{fig:comp_on_psv} show the qualitative comparisons of our method with existing methods on Places2, CelebA and {\color{black}{Paris StreetView}} respectively. As shown in these figures, PatchMatch can synthesize smooth textures, but it can not synthesize semantically correct content when the holes are crossing foreground objects. PConv and LBAM can generate plausible structures, but artifacts still exist when the holes are relatively large. 
CA does not work well with the irregular masks since it is not specially designed for irregular masks.
EdgeConnect shows the potential of generating highly structural images with the hallucinated edges from its first stage, but it fails to generate correct structure when the hallucinated edges contain errors. 
GConv is able to generate both smooth and plausible textures, but the results still show unpleasant flaws. 
Compared with the results of these baselines, details of our results are enhanced and the generated edges are kept distinct and smooth, even though the damaged contents are relatively large and complex. {\color{black}{We contribute this to the characteristic of our method,
which utilizes the mask awareness and refines the result progressively in a cascade way. With mask awareness, our model can learn to generate better features representations to denote the image content for decoding. Besides, at the decoding phase, PN is leverage to eliminate ``covariant shift'' when normalizing the features. Therefore, the features of hole regions at decoder is as effective and homogeneous as the non-hole regions. Further combining refinement decoders, the results become more plausible.}}
}

\begin{figure}[!h]
\centering
\includegraphics[scale=0.6]{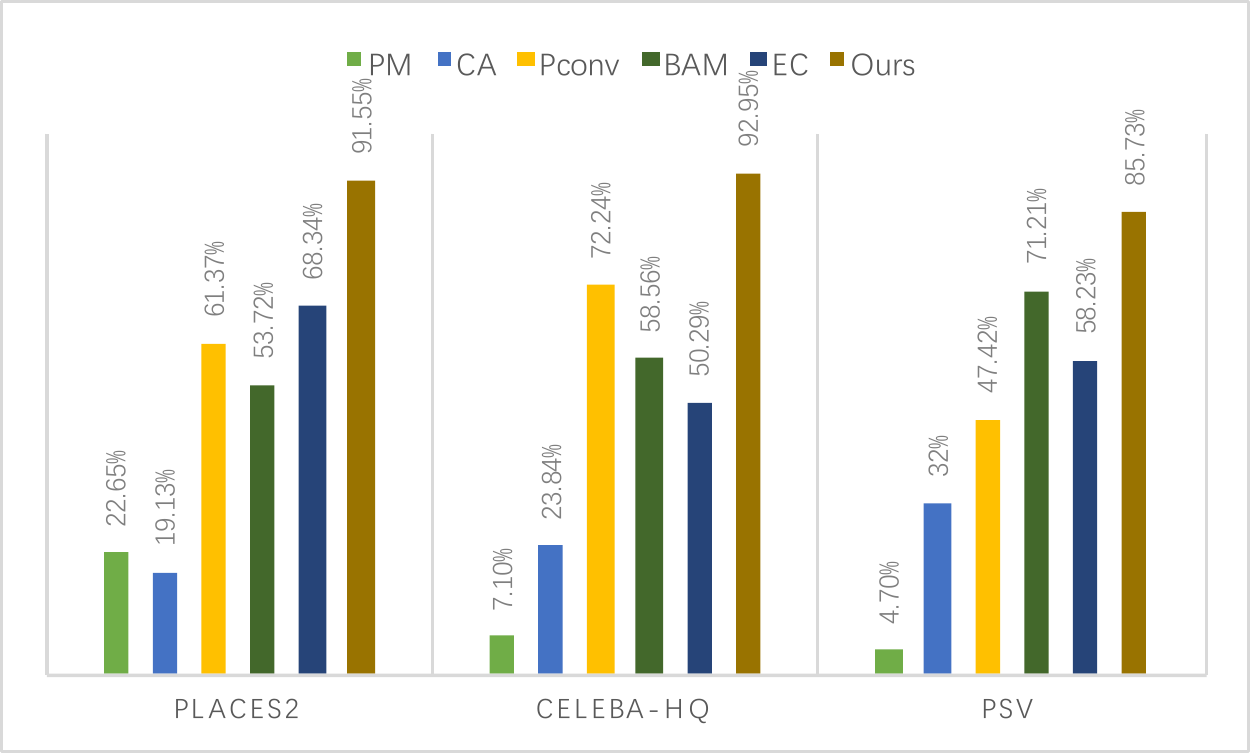}
\caption{User study results. The value indicates the percentage of being selected as the better one.}
\label{fig:user_study}
\end{figure}
\subsection{User study}
There is a gap between the evaluation metrics and human perception,
therefore we conduct user study on these datasets for a more credible conclusion. We invite 20 volunteers with image processing background to participate in the experiments.
For each dataset, 100 images with random masks are randomly sampled from the test set. We perform pairwise A/B tests. Each time, a pair of generated images by different methods are shown to the volunteers. Without telling the mask information or the original ground truth image, the volunteers are asked to choose the more visually plausible and natural one from the two images. The results are shown in Figure \ref{fig:user_study}. As can be seen, most of time volunteers prefer our results at all these datasets. 

\begin{figure}[!h]
\begin{center}
\begin{tabular}{ccc}
\centering
\includegraphics[width=0.3\columnwidth]{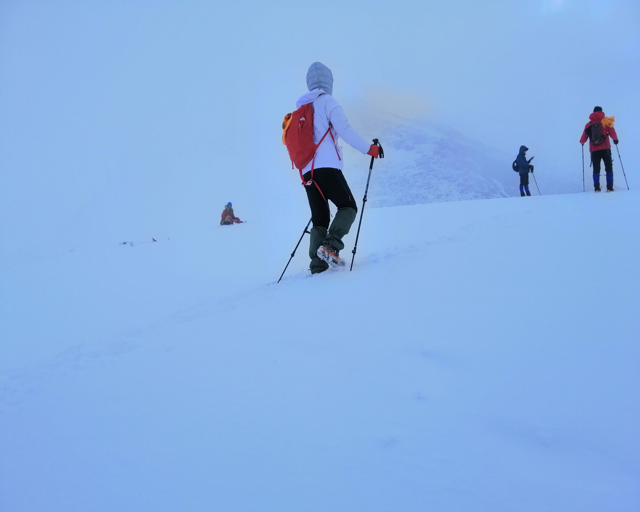}&
\includegraphics[width=0.3\columnwidth]{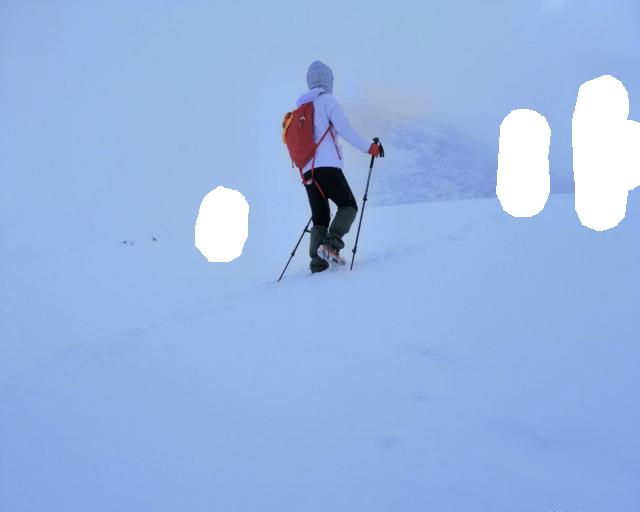}&\includegraphics[width=0.3\columnwidth]{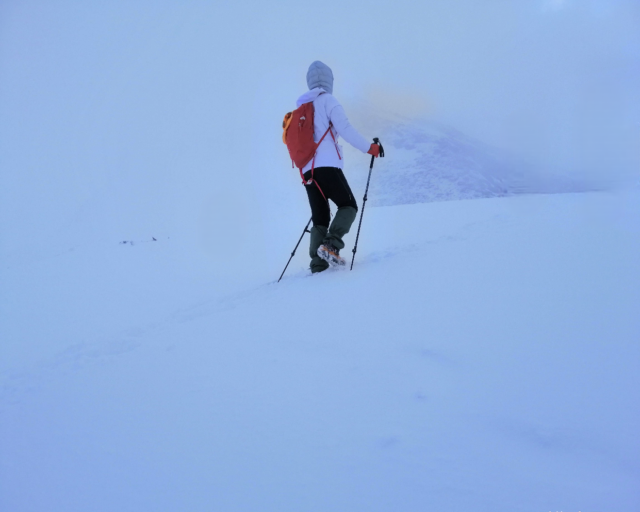}\\
\includegraphics[width=0.3\columnwidth]{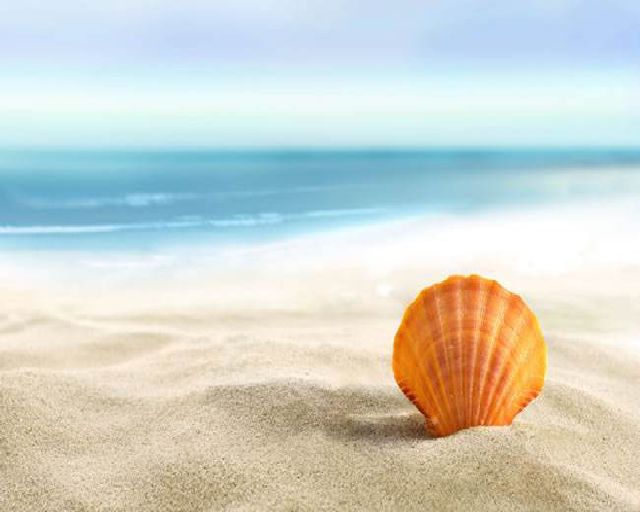}&
\includegraphics[width=0.3\columnwidth]{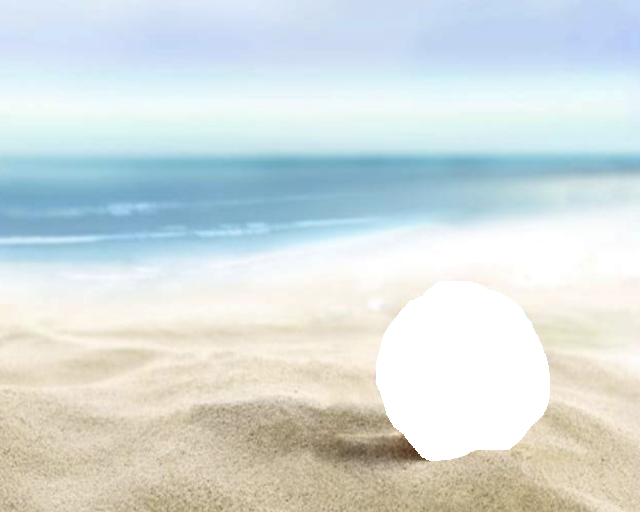}&\includegraphics[width=0.3\columnwidth]{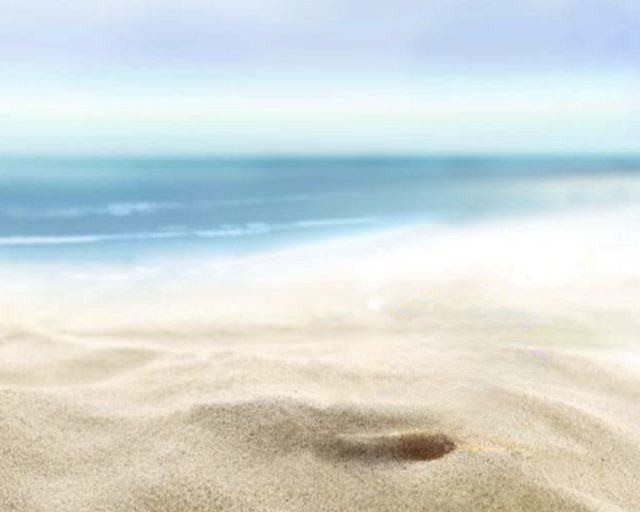}\\
\end{tabular}
\end{center}
\caption{Examples of object removal using our framework. From left to right: original images, unwanted objects removed with masks, generated images. Best viewed with zoom-in.}
\label{fig:real_case}
\end{figure}
\subsection{Application Showcase}
To show the generalization ability, we apply our model trained on Places2 to natural images downloaded from web.
Figure \ref{fig:real_case} shows some examples of object removal results using our framework which is one of the most important real use cases of image inpainting. Our model shows great generalization capability. 

\begin{figure}[h]
\begin{center}
\setlength{\tabcolsep}{0.5mm}
\begin{tabular}{cccc}
\centering
\includegraphics[width=0.23\columnwidth, height=0.2\columnwidth]{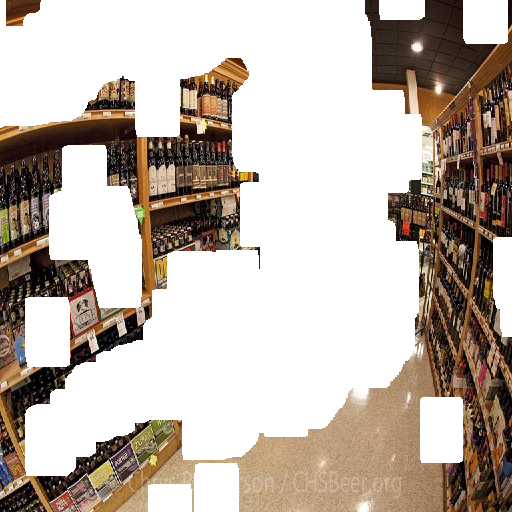}&\includegraphics[width=0.23\columnwidth, height=0.2\columnwidth]{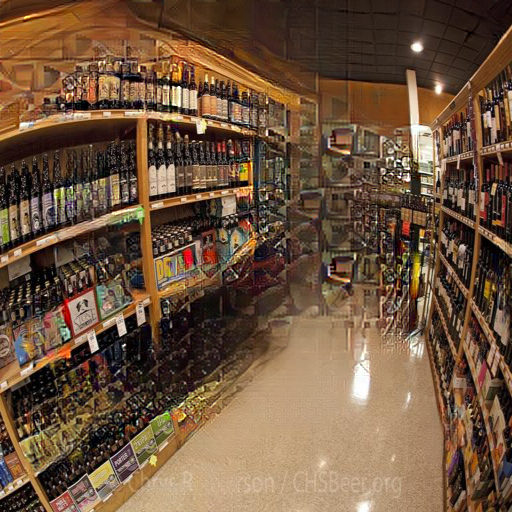}&\includegraphics[width=0.23\columnwidth, height=0.2\columnwidth]{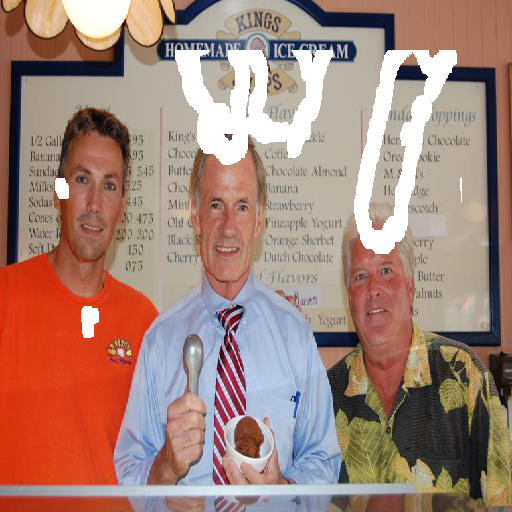}&\includegraphics[width=0.23\columnwidth, height=0.2\columnwidth]{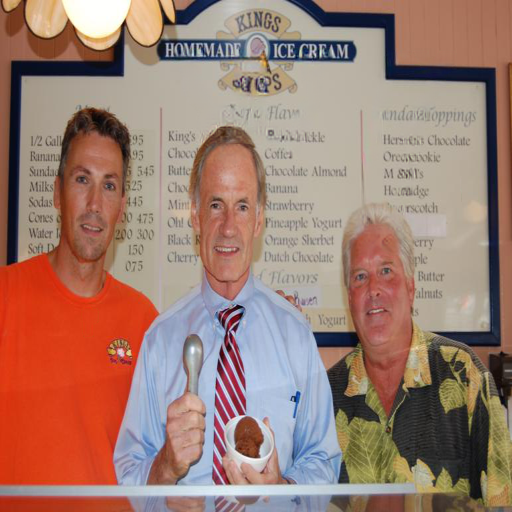}\\
\end{tabular}
\end{center}
\caption{Demonstration of two failure cases on Places2 dataset. Best viewed with zoom-in.}
\label{fig:casestudy}
\end{figure}

{\color{black}{\subsection{Discussion on Failure Cases}
To make clear the limitations of our models, we conduct case studies on Places2. Several failure cases are presented in Figure \ref{fig:casestudy}. From these results, we find that our model works not so well when large missing region is encountered. In such case, the model trends toward filling in the holes with smooth content information according to its surroundings. It is because context information is very limited and a general inpainting model cannot well hallucinate the lost region. Besides, when very special information is lost, such as texts and logos, the model also fails due to its lack of knowledge. Interestingly, our model trained on CelebA can produce plausible results when a large potion of face region is missing. This is because the model is specially trained for face scenario. Conditioned on the surroundings, it is not so difficult for a specifically well trained model to generate faces.}}

\section{Conclusion}
In this paper, we propose a novel framework which can fill arbitrary missing regions for images in a cascaded refinement with mask awareness fashion. The novel MADF module generates convolution kernels adaptively to corresponding regions in the mask for all convolution windows, thus the extracted feature can get awareness of validity of each point in a convolution window and be more representative for decoding. Intuition of mask awareness also motivates us to adopt point-wise normalization, which is proven to be powerful.
Besides, in contrast to two-stage methods that use hand-crafted clues, our model is of cascaded refinement design and uses high-level multi-scale features as the guidance for pixel generation from coarse to fine. 
Extensive experiments in various scenes verify that our method outperforms the current state-of-the-art methods in both subjective visual quality and objective quantitative measurements. Codes and pretrained models are released: \url{https://github.com/MADF-inpainting/Pytorch-MADF}.


%

\ifCLASSOPTIONcaptionsoff
  \newpage
\fi



%

\bibliographystyle{IEEEtran}
\bibliography{egbib}

%
\iftrue
\vspace{-40px}
\begin{IEEEbiography}[{\includegraphics[width=0.8in,height=1in,clip,keepaspectratio]{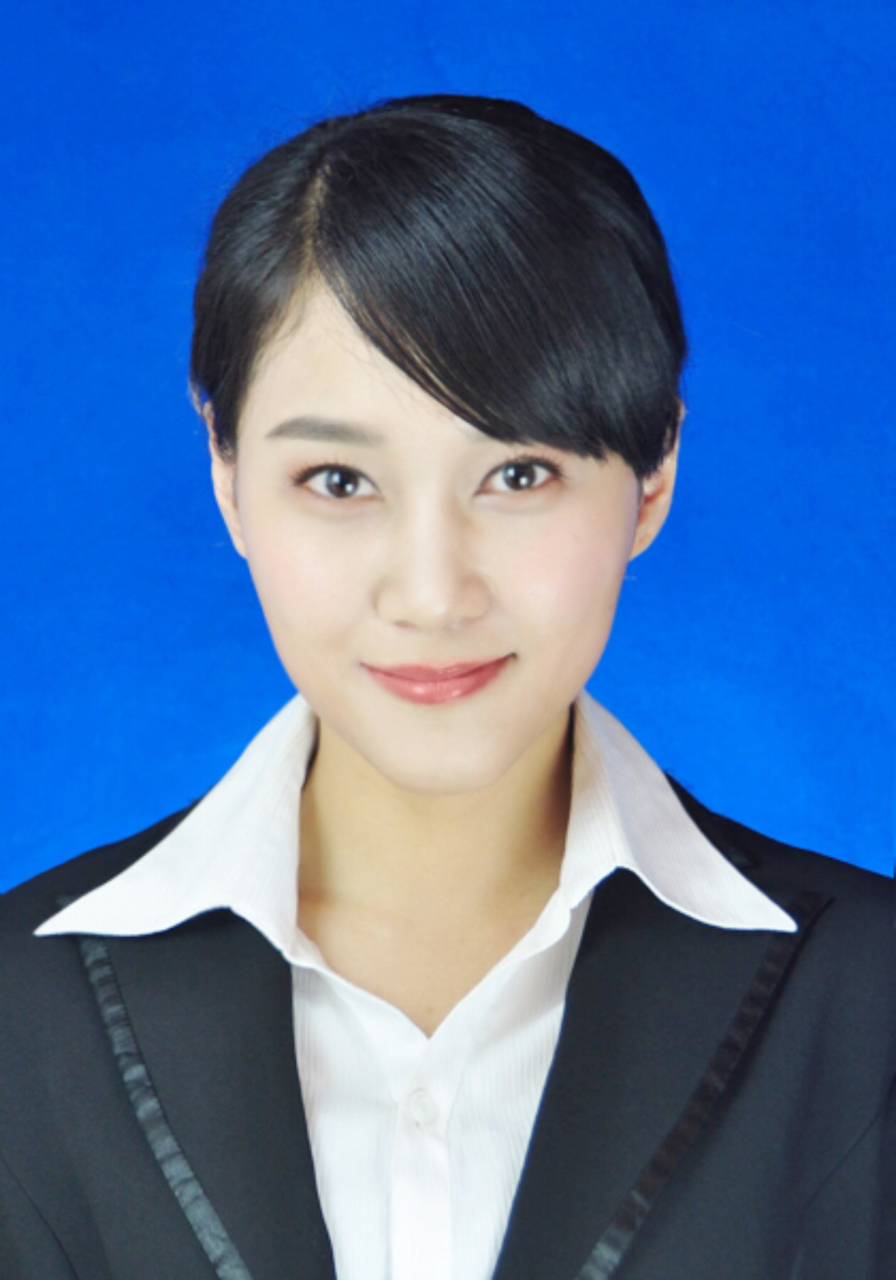}}]{Manyu Zhu}
Manyu Zhu is a senior research and development engineer at Department of Computer Vision (VIS) Technology, Baidu Inc. She received the M.S. degree in pattern recognition and intelligent system from Huazhong University of Science and Technology. Her research interests focus on computer vision based on deep learning, image processing and multimodal deep learning.
\end{IEEEbiography}

\vspace{-60px}
\begin{IEEEbiography}[{\includegraphics[width=0.8in,height=1in,clip,keepaspectratio]{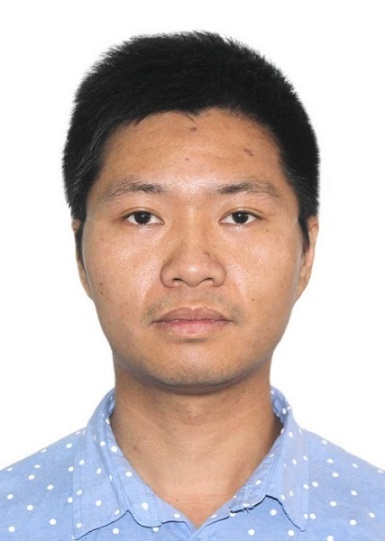}}]{Dongliang He}
Dongliang He is a senior research and development engineer at Department of Computer Vision (VIS) Technology, Baidu Inc. He received Bachelor and  Ph.D. degree in computer science from University of Science and Technology of China in 2012 and 2017, respectively. His research interests focus on computer vision, deep learning and multi-media processing. 
\end{IEEEbiography}

\vspace{-60px}
\begin{IEEEbiography}[{\includegraphics[width=0.8in,height=1in,clip,keepaspectratio]{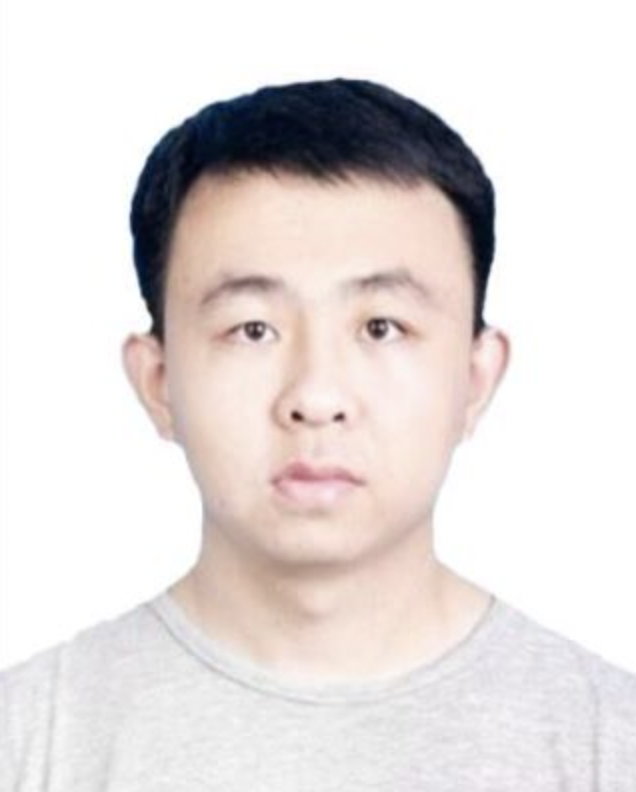}}]{Xin Li}
Lixin is a senior research and development engineer at Department of Computer Vision (VIS) Technology, Baidu Inc. He received the M.S. degree in EE from TsingHua University. His research interests focus on computer vision based on deep learning, image processing.
\end{IEEEbiography}

\vspace{-60px}
\begin{IEEEbiography}[{\includegraphics[width=0.8in,height=1in,clip,keepaspectratio]{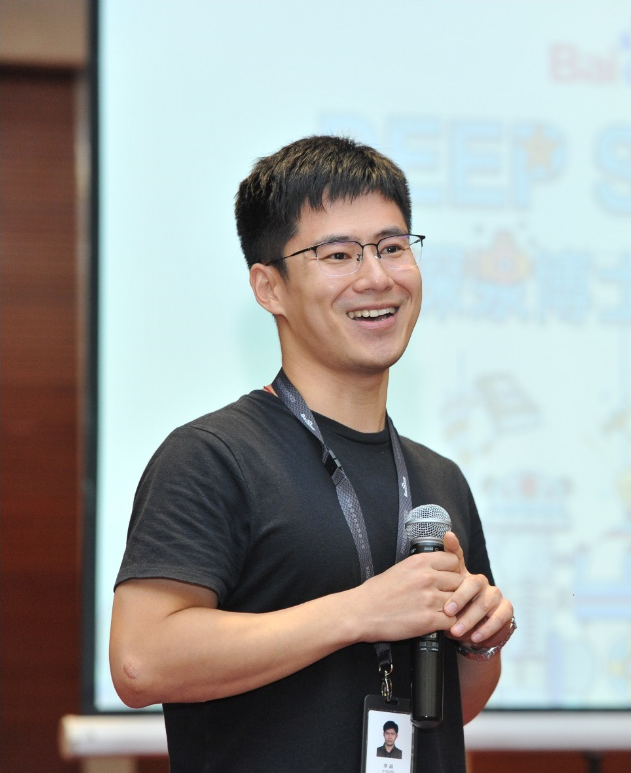}}]{Chao Li} is a senior researcher and engineer at Department of Computer Vision (VIS) Technology, Baidu Inc. He received his Ph.D. degree in The University of Queensland, Australia, in 2018. His research interests include Computer Vision, Multimedia Search and Artificial Intelligence. 
\end{IEEEbiography}

\vspace{-60px}
\begin{IEEEbiography}[{\includegraphics[width=0.8in,height=1in,clip,keepaspectratio]{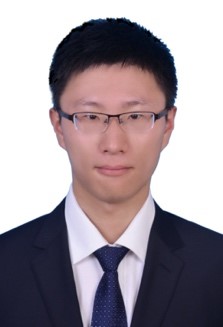}}]{Fu Li}
Fu Li is a senior R\&D engineer at Department of Computer Vision Technology (VIS), Baidu Inc. He received M.E. degree from Dalian University of Science and Technology, China in 2015. His research interests focus on video understanding and generative adversarial networks.
\end{IEEEbiography}

\vspace{-60px}
\begin{IEEEbiography}[{\includegraphics[width=0.8in,height=1in,clip,keepaspectratio]{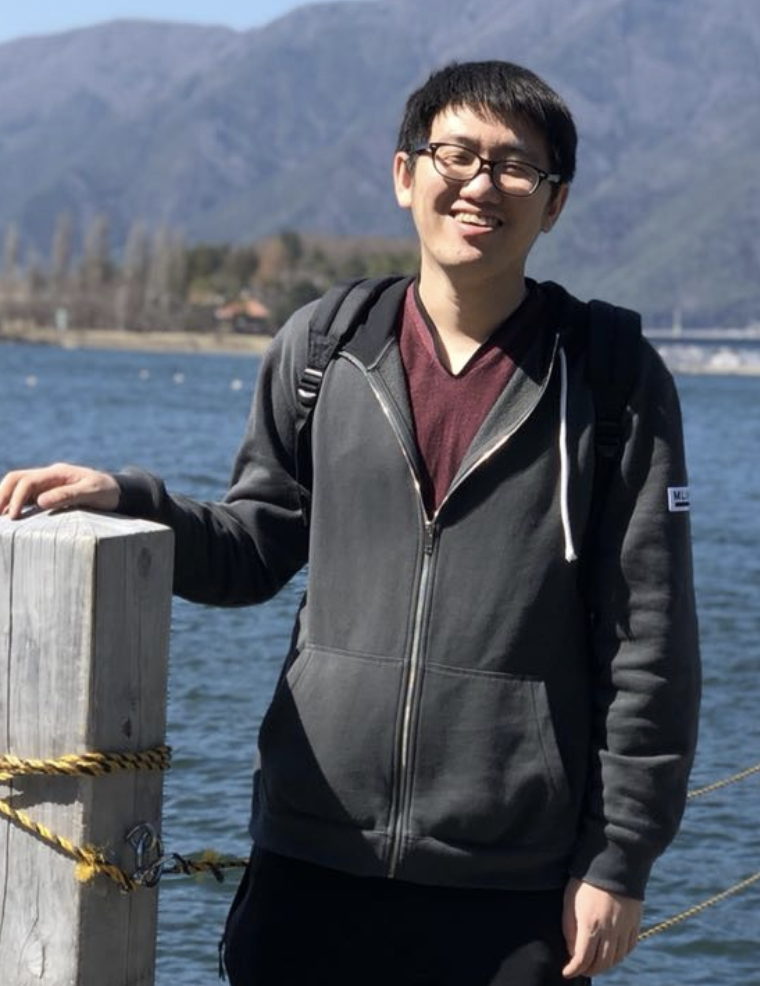}}]{Xiao Liu}
Xiao Liu is a researcher in the Tomorrow Advancing Life Education Group (TAL). He received the PhD degree in Computer Science from Zhejiang University, China, in 2015. He previously worked for Baidu from 2015 to 2019.
His research interests include computer vision and learning system.
\end{IEEEbiography}


\vspace{-60px}
\begin{IEEEbiography}[{\includegraphics[width=0.8in,height=1in,clip,keepaspectratio]{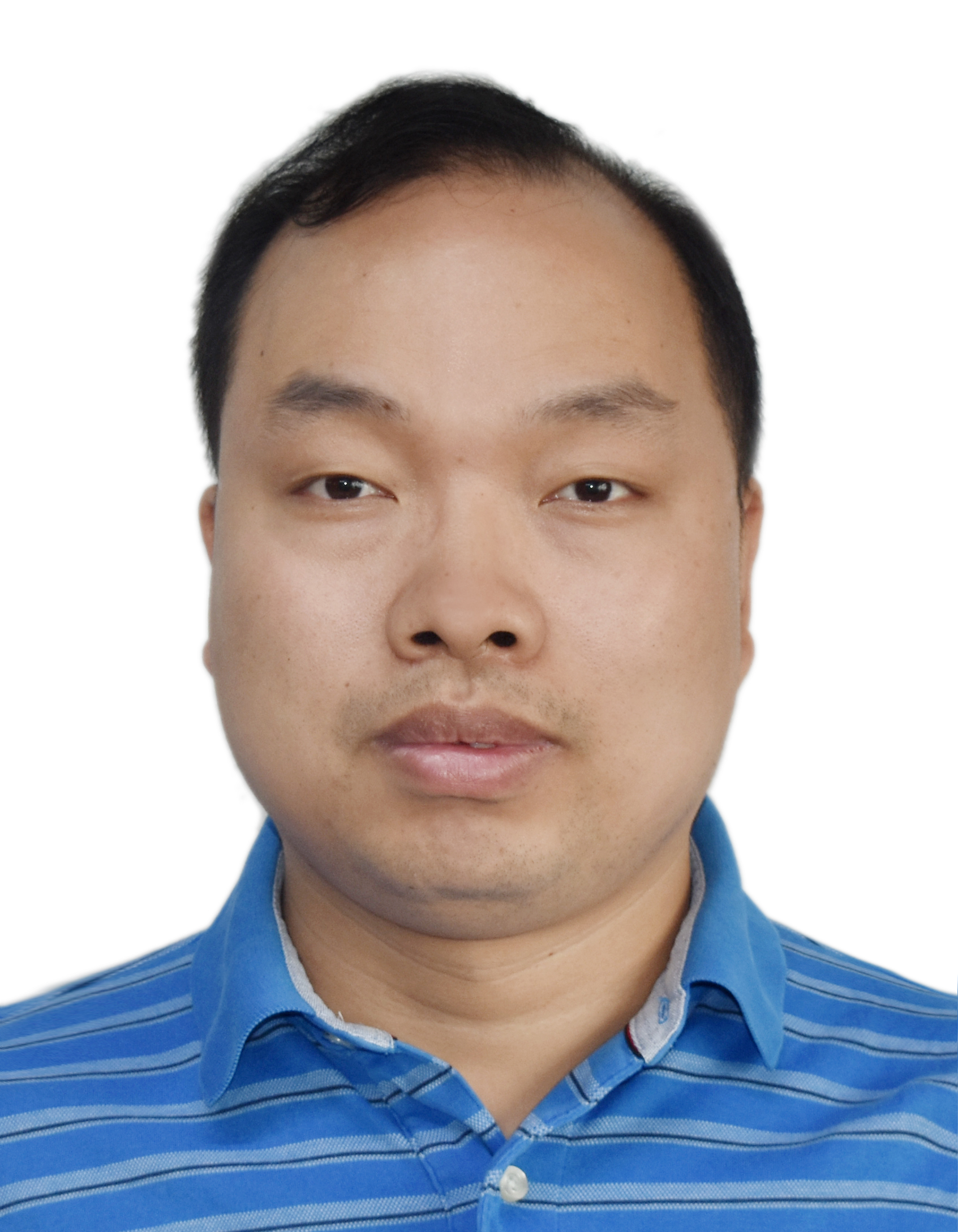}}]{Errui Ding}
Errui Ding received Ph.D degree in 2008 from Xidian University, and currently is the director of Computer Vision Technology Department (VIS) of Baidu Inc.  In recent years, he has published tens of papers on top-tier conferences and was awarded Best Paper Runner-up by ICDAR 2019. He also co-organized two competition tracks, i.e., ArT and CSVT, in ICDAR 2019 and the 2nd Workshop on Learning from Imperfect Data in CVPR 2020. As a member, he serves in special committee of China Society of Image and Graphics.
\end{IEEEbiography}

\vspace{-40px}
\begin{IEEEbiography}[{\includegraphics[width=0.8in,height=1in,clip,keepaspectratio]{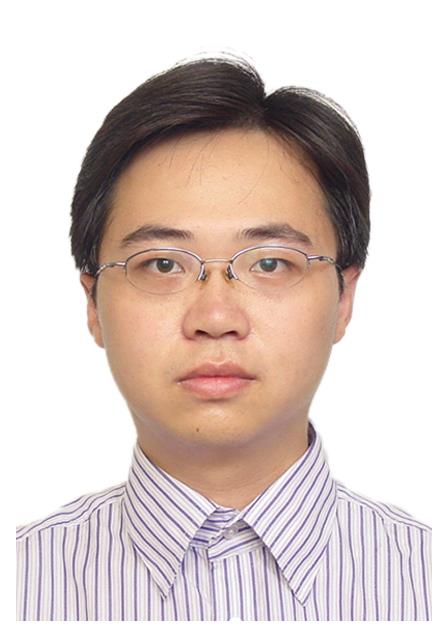}}]{Zhaoxiang Zhang}
Zhaoxiang Zhang is currently a Professor in the Center for Research on Intelligent Perception and Computing, Institute of Automation, Chinese Academy of Sciences (CASIA), Beijing, China. He received his bachelor's degree from the University of Science and Technology of China (USTC) in 2004 and he received his Ph.D. degree in 2009 from CAS. 
His major research interests include pattern recognition, computer vision, machine learning and bio-inspired visual computing. 
He has won the best paper awards in several conferences and championships in international competitions. He has served as the area chair, senior PC of many international conferences like CVPR, ICCV, AAAI, IJCAI. He is now serving as the associate editor of IEEE T-CSVT, Pattern Recognition and Neurocomputing.
\end{IEEEbiography}



\fi



\end{document}